%% file: main.tex
\documentclass[10pt,journal,compsoc]{IEEEtran}
\ifCLASSOPTIONcompsoc
  \usepackage[nocompress]{cite}
\else
  \usepackage{cite}
\fi
\ifCLASSINFOpdf
\else
\fi

\usepackage{url}            
\usepackage{nicefrac}       
\usepackage{microtype}      
\usepackage{floatrow}
\newfloatcommand{capbtabbox}{table}[][\FBwidth]
\usepackage{xcolor}

\usepackage[utf8]{inputenc} 
\usepackage[T1]{fontenc}    
\usepackage{url}            
\usepackage{amsfonts}       
\usepackage{nicefrac}       
\usepackage{microtype}      
\usepackage{xcolor}         
\usepackage{multirow} 

\usepackage{algorithm}
\usepackage{algpseudocode}

\usepackage{graphicx}
\graphicspath{{./}{../}}
\usepackage{booktabs}
\usepackage{times}
\usepackage{epsfig}
\usepackage{amsmath}
\usepackage{amssymb}
\usepackage{enumitem}
\usepackage[font=small]{caption}
\usepackage{subcaption}
\usepackage{array}
\usepackage[bookmarks=false]{hyperref}  
\hypersetup{                           
    colorlinks = true,                 
    citecolor  = blue,                 
    linkcolor  = blue,                 
    urlcolor   = blue,                 
} 
\usepackage{cleveref}
\Crefformat{figure}{Fig.~#2#1#3}                           
\Crefname{subfigure}{Fig.}{Figs.}
\Crefname{figure}{Fig.}{Figs.}
\Crefformat{table}{TABLE~#2#1#3}                           
\usepackage[skip=5pt]{caption}            
\begin{document}

\twocolumn

\title{\huge
    Boosting Fidelity for Pre-Trained-Diffusion-Based Low-Light Image Enhancement via Condition Refinement
}

\author{
    Xiaogang Xu, Jian Wang, Yunfan Lu, Ruihang Chu, Ruixing Wang, Jiafei Wu, \\  Bei Yu and Liang Lin (IEEE Fellow)
    \thanks{Xiaogang Xu and Bei Yu are with the Department of Computer Science and Engineering, The Chinese University of Hong Kong, E-mail: xiaogangxu00@gmail.com, byu@cse.cuhk.edu.hk}
    \thanks{Jian Wang is with Snap Research, E-mail:  jwang4@snapchat.com}
    \thanks{Yunfan Lu is with HKUST (GZ), E-mail:  ylu066@connect.hkust-gz.edu.cn}
    \thanks{Ruihang Chu is Alibaba Tongyi Lab, E-mail: ruihangchu@gmail.com}
    \thanks{Ruixing Wang is with the camera group of DJI, E-mail: ruixingw@hustunique.com}
    \thanks{Jiafei Wu is with The University of Hong Kong, E-mail: jcjiafeiwu@gmail.com}
    \thanks{Liang Lin is with Sun Yat-Sen University, E-mail: linlng@mail.sysu.edu.cn}
}

\markboth{submission to IEEE Transactions on Pattern Analysis and Machine Intelligence}
{Shell \MakeLowercase{\textit{et al.}}: Bare Demo of IEEEtran.cls for Computer Society Journals}

\IEEEtitleabstractindextext{%
\begin{abstract}
Diffusion-based methods, leveraging pre-trained large models like Stable Diffusion via ControlNet, have achieved remarkable performance in several low-level vision tasks. 
However, Pre-Trained Diffusion-Based (PTDB) methods often sacrifice content fidelity to attain higher perceptual realism.
This issue is exacerbated in low-light scenarios, where severely degraded information caused by the darkness limits effective control.
We identify two primary causes of fidelity loss: the absence of suitable conditional latent modeling and the lack of bidirectional interaction between the conditional latent and noisy latent in the diffusion process. To address this, we propose a novel optimization strategy for conditioning in pre-trained diffusion models, enhancing fidelity while preserving realism and aesthetics. Our method introduces a mechanism to recover spatial details lost during VAE encoding, i.e., a latent refinement pipeline incorporating generative priors. Additionally, the refined latent condition interacts dynamically with the noisy latent, leading to improved restoration performance.
Our approach is plug-and-play, seamlessly integrating into existing diffusion networks to provide more effective control. Extensive experiments demonstrate significant fidelity improvements in PTDB methods.
\end{abstract}

}

\maketitle
\thispagestyle{plain}
\pagestyle{plain}

\IEEEdisplaynontitleabstractindextext

%
\IEEEpeerreviewmaketitle

\input{sec/1_intro}

\input{sec/2_related_work}

\input{sec/3_method}
\input{sec/4_experiment}

\input{sec/5_conclusion}



%


\bibliographystyle{IEEEtran}
\bibliography{egbib}

%

\vspace{-.4in}
\begin{IEEEbiography}
    [{\includegraphics[height=1.15in,clip,keepaspectratio]{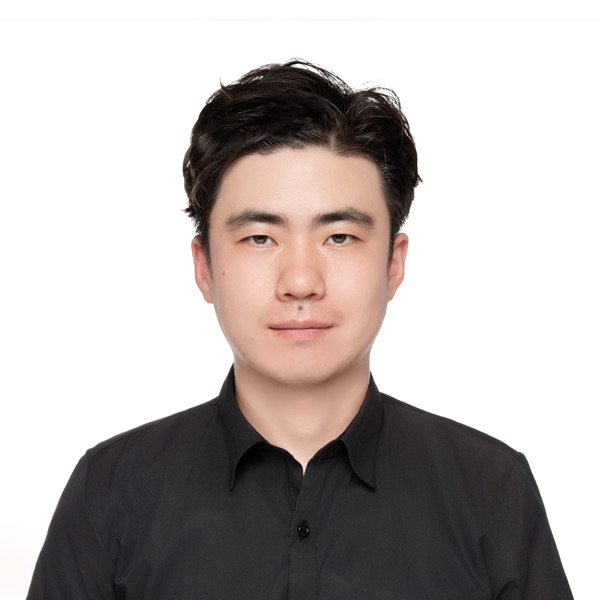}}]
    {Xiaogang Xu}
    is a postdoc research fellow in the Chinese University of Hong Kong. He received his Ph.D. degree from CUHK in 2022 and bachelor degree from Zhejiang University in 2018. In 2023, he is a research scientist in Zhejiang Lab and meanwhile a ZJU100 Young Professor at ZJU.
	He obtained the Hong Kong PhD Fellowship in 2018. He serves as a reviewer for CVPR, ICCV, ECCV, Neurips, ICLR, TPAMI, TIP, IJCV, etc. His research interest includes deep learning, computational photography, AIGC, large models.
\end{IEEEbiography}

\vspace{-.4in}
\begin{IEEEbiography}
	[{\includegraphics[height=1.1in,clip,keepaspectratio]{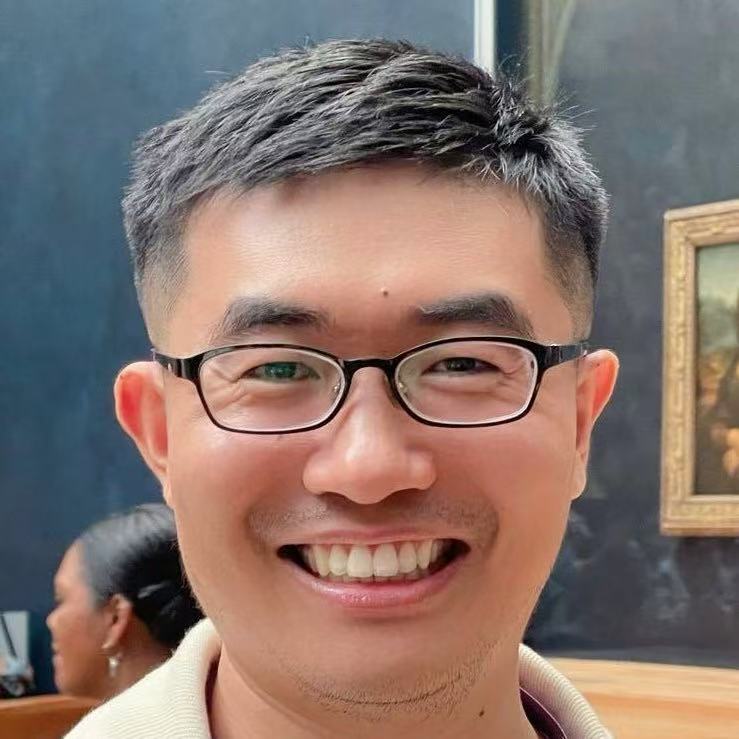}}]
	{Jian Wang} is a Staff Research Scientist at Snap Inc., focusing on computational photography and imaging. He has published in top-tier venues such as CVPR, MobiCom, and SIGGRAPH, and has contributed numerous features to production. He has received the best paper award from SIGGRAPH Asia, 2024, and the 4th IEEE International Workshop on Computational Cameras and Displays, and the best poster award from IEEE Conference on Computational Photography 2022. He has served as an Area Chair for CVPR, NeurIPS, ICLR, ICML, etc. Jian holds a Ph.D. from Carnegie Mellon University.
\end{IEEEbiography}

\vspace{-.4in}
\begin{IEEEbiography}
	[{\includegraphics[height=1.1in,clip,keepaspectratio]{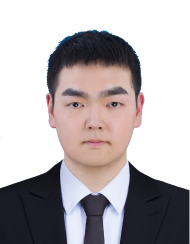}}]
	{Yunfan Lu} received the B.S. degree in Computer Science from Nanjing University of Science and Technology and the M.S. degree in Computer Science from the University of Chinese Academy of Sciences. He is currently a final-year Ph.D. student at the Hong Kong University of Science and Technology (Guangzhou). His research interests include computational imaging, knowledge mining from image and video data, and event camera technology. He has served as a reviewer for major international conferences and journals, including CVPR, ECCV, T-PAMI, and IJCV.
\end{IEEEbiography}

\vspace{-.4in}
\begin{IEEEbiography}
	[{\includegraphics[height=1.1in,clip,keepaspectratio]{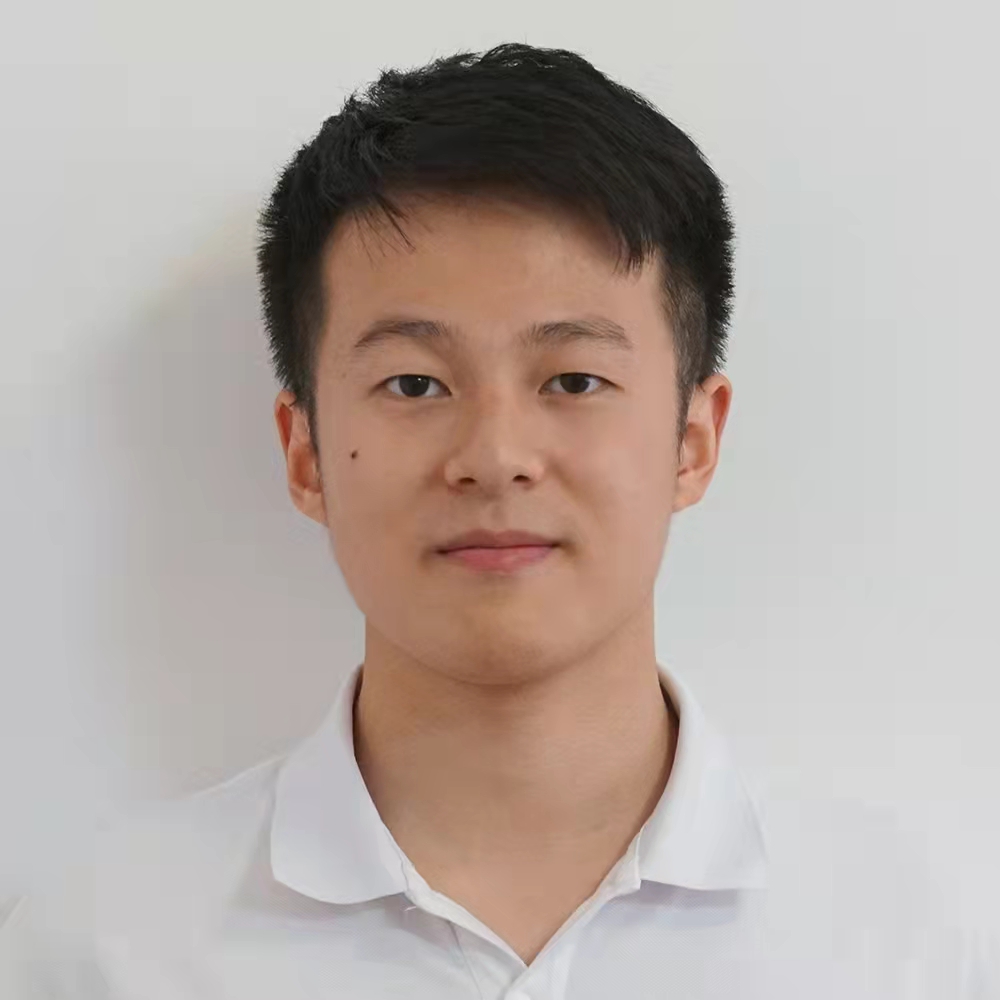}}]
	{Ruihang Chu} received the Ph.D. degree in computer science and engineering from The Chinese University of Hong Kong, Hong Kong, in 2024. In 2017 and 2020, he received the B.E. degree and M.E degree in mechanical engineering and automation from Beihang University, Beijing, respectively. His research interests include visual generation and multi-modal large language models.
\end{IEEEbiography}

\vspace{-.4in}
\begin{IEEEbiography}
	[{\includegraphics[height=1.15in,clip,keepaspectratio]{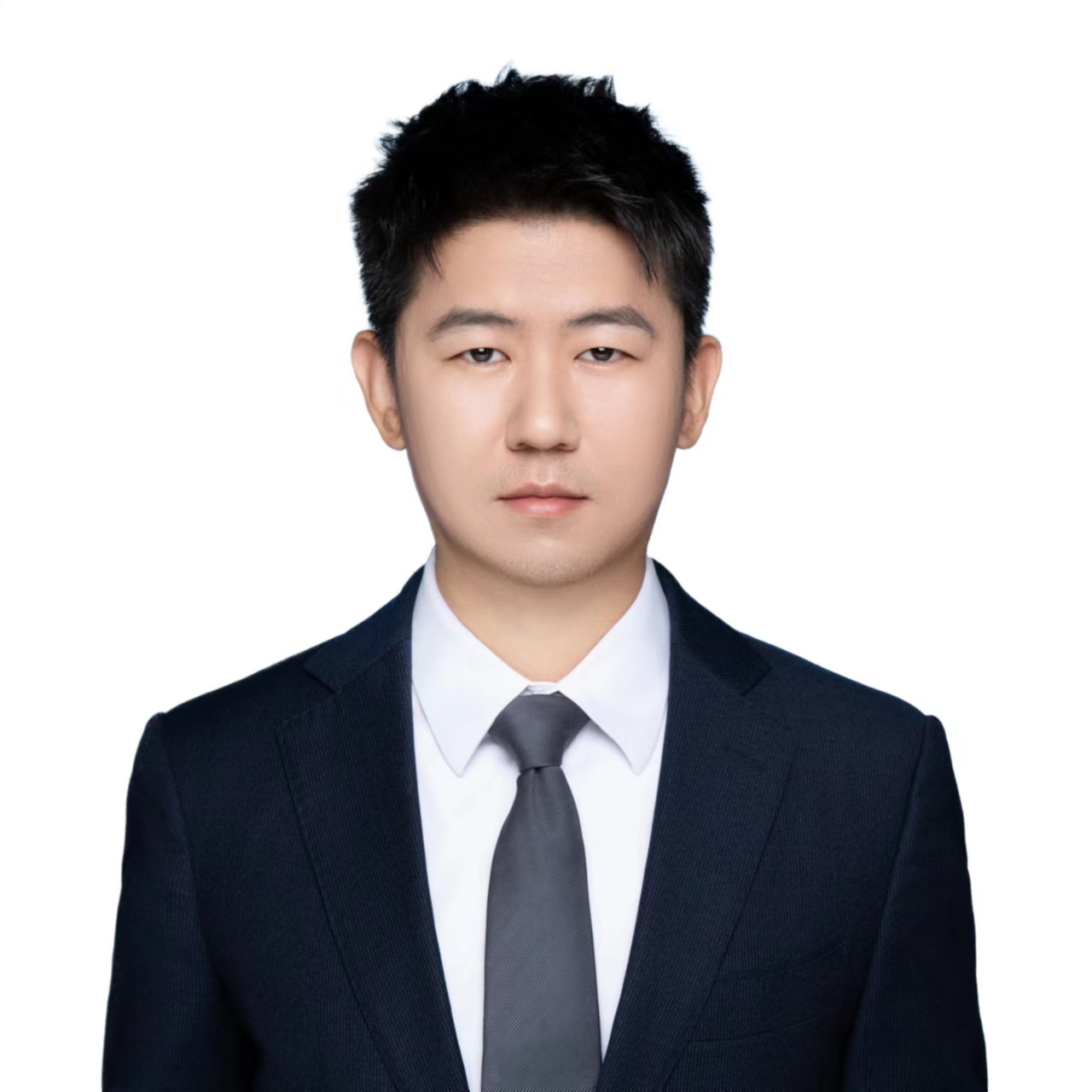}}]
	{Ruixing Wang} is currently at camera group of DJI. He received the B.S. degree from Huazhong University of Science and Technology in 2016, and the Ph. D. degree from the Chinese University of Hong Kong, in 2021. Before joining in DJI, he was a principal engineer in Honor Device Co., Ltd. He serves as a reviewer for CVPR, ICCV, ECCV, Neurips, ICML, ICLR, AAAI, WACV, ACCV, TPAMI, IJCV, etc. His research interests include computational photography and image processing. 
\end{IEEEbiography}

\vspace{-.4in}
\begin{IEEEbiography}
	[{\includegraphics[height=1.25in,clip,keepaspectratio]{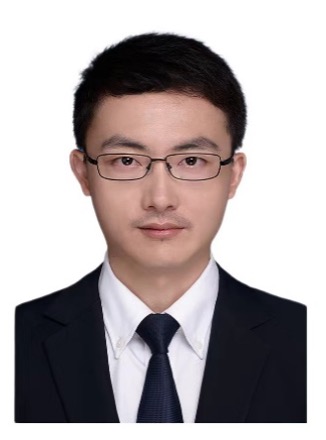}}]
	{Jiafei Wu}
	received the B.S. degree from JXUFE in 2010, the M.S. degree and Ph.D. degree from the University of Hong Kong in 2012 and 2017, respectively. He has been a senior engineer, manager and deputy director from 2018 to 2023 in SenseTime. He is currently with the Zhejiang Lab. His research interests include deep learning, trustworthy AI, embedded system, and computational intelligence.
\end{IEEEbiography}

\vspace{-.4in}
\begin{IEEEbiography}
    [{\includegraphics[height=1.26in,clip,keepaspectratio]{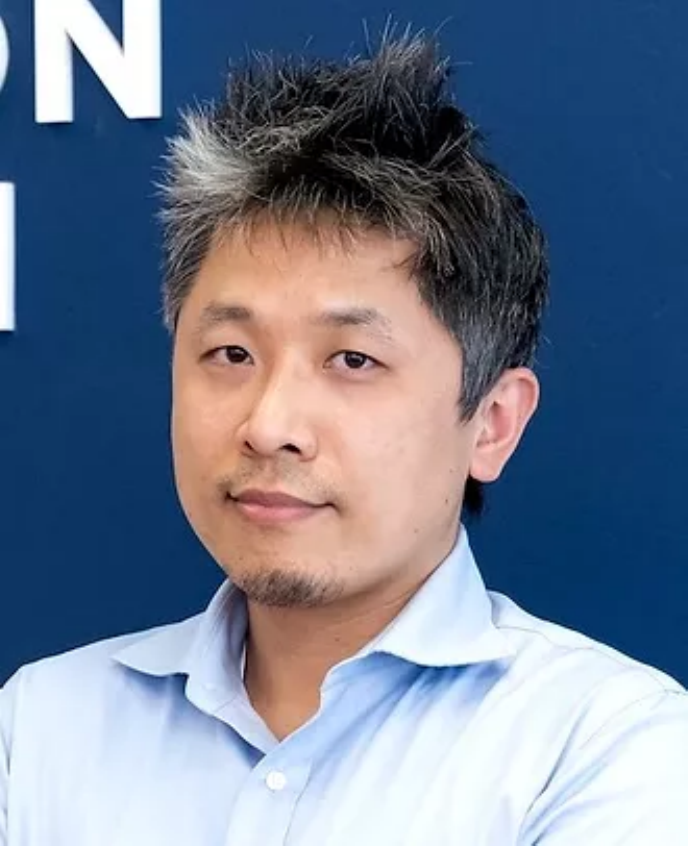}}]
    {Bei Yu}
    (M'15-SM'22)
    received the Ph.D.~degree from The University of Texas at Austin in 2014.
    He is currently a Professor in the Department of Computer Science and Engineering, The Chinese University of Hong Kong.
    He has served as TPC Chair of ACM/IEEE Workshop on Machine Learning for CAD, and in many journal editorial boards and conference committees.
    He received eleven Best Paper Awards from ICCAD 2024 \& 2021 \& 2013,
    IEEE TSM 2022, DATE 2022, ASPDAC 2021 \& 2012, ICTAI 2019, Integration, the VLSI Journal in 2018,
    ISPD 2017, SPIE Advanced Lithography Conference 2016, and six ICCAD/ISPD contest awards.
\end{IEEEbiography}

\vspace{-.4in}
\begin{IEEEbiography}
	[{\includegraphics[height=1.25in,clip,keepaspectratio]{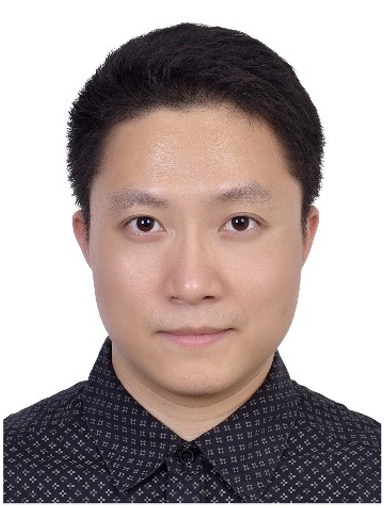}}]
	{Liang Lin} (M'09, SM'15, F'24) is a Full Professor of computer science at Sun Yat-sen University. His research focuses on new models, algorithms and systems for intelligent processing and understanding of multimodal data. He has authored or co-authored more than 400 papers in leading academic journals and conferences, and his papers have been cited by more than 45,000 times. He is an associate editor of IEEE Trans. Multimedia and IEEE Trans. Neural Networks and Learning Systems, and served as Area Chairs for numerous conferences such as CVPR, ICCV, SIGKDD and NeurIPS. He is the recipient of numerous awards and honors including CCF-ACM Award for Artificial Intelligence, Wu Wen-Jun Artificial Intelligence Award, the First Prize of China Society of Image and Graphics, ACL Outstanding Paper Award in 2025, ICCV Best Paper Nomination in 2019, Annual Best Paper Award by Pattern Recognition (Elsevier) in 2018, Best Paper Dimond Award in IEEE ICME 2017, Google Faculty Award in 2012. His supervised PhD students received ACM China Doctoral Dissertation Award, CCF Best Doctoral Dissertation and CAAI Best Doctoral Dissertation. He is a Fellow of IET, IAPR, and IEEE.
\end{IEEEbiography}

\end{document}

%% file: sec/1_intro.tex
\section{Introduction}
\label{sec:intro}

Pre-Trained-Diffusion-Based (PTDB) methods~\cite{rombach2022high,flux2024} have been leveraged for low-level tasks by using degraded inputs as conditions to guide the generation process of a pre-trained large text-to-image model.
They normally employ conditional frameworks like ControlNet~\cite{zhang2023adding}.
Note that the reference-based metrics (which measure fidelity, e.g., PSNR, SSIM, LPIPS) of PTDB methods are often lower than those of traditional restoration methods, as the core mechanism of PTDB is generation rather than pixel-wise reconstruction.
However, compared to traditional restoration networks, PTDB models generate more aesthetically pleasing and visually appealing results by leveraging their pre-trained knowledge to synthesize rich high-resolution details. \textit{Consequently, many researchers and companies have devoted increasing attention to this field, with the primary goal of enhancing fidelity while retaining the inherent strengths of these models}.

\begin{figure}[t]
    \begin{center}
        \includegraphics[width=1\linewidth]{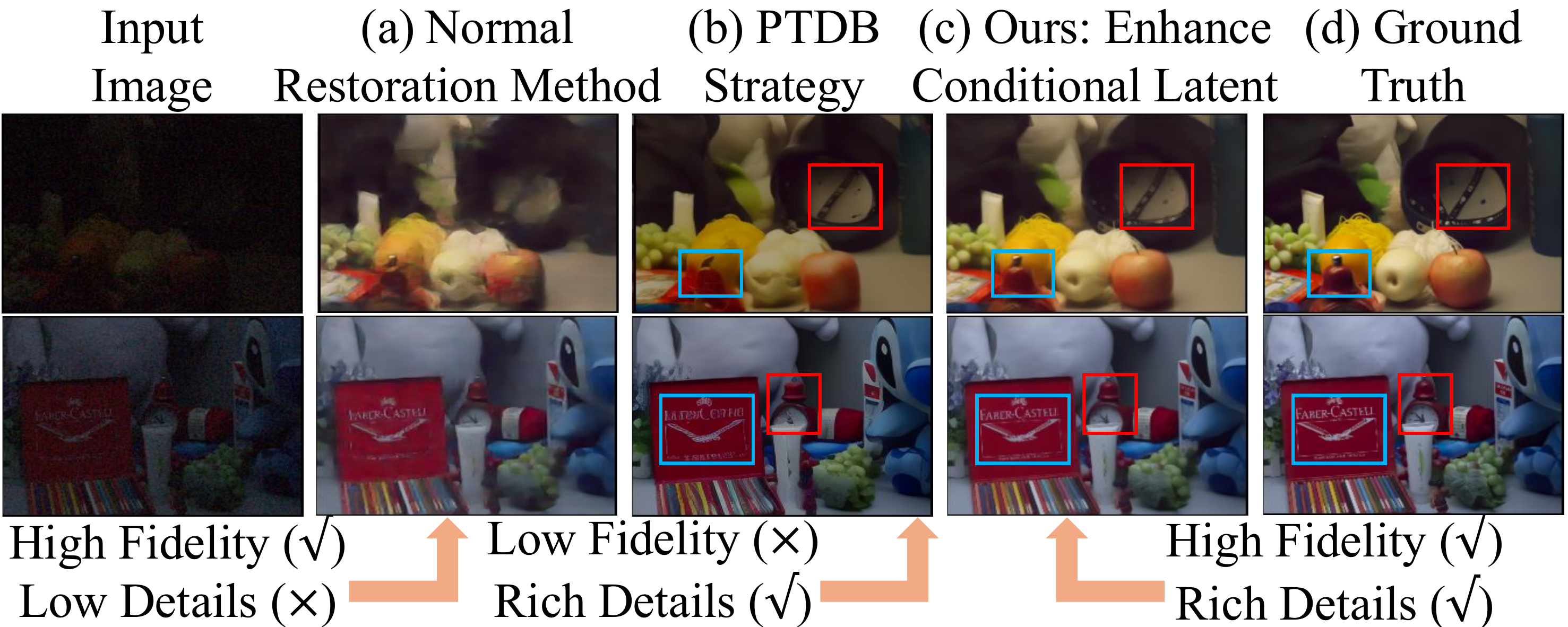}
    \end{center}
    \vspace{-0.15in}
    \caption{
(a) Traditional restoration-only methods (SNR-aware network~\cite{xu2022snr} in the top and Diff-L~\cite{jiang2023low} that is trained from scratch) achieve high fidelity, while users may not accept these noisy results in practice.
They require improvements in detail and aesthetic quality. 
(b) In contrast, PTDB methods (e.g., DiffBIR~\cite{lin2024diffbir} here) leverage pre-trained diffusion models to generate detailed, sharp, and clean images, but often at the loss of fidelity (distortion). Fidelity is critical for some regions, e.g., letters. 
(c) Unlike prior works that focus on enhancing control structures for diffusion models, our approach directly refines the conditional latents and their interaction with noisy latents at various steps. This allows us to preserve advantages of pre-trained diffusion models while improving fidelity. 
Our method is plug-and-play for all PTDB approaches.
	}
	\label{fig:teaser}
\end{figure}

On the other hand, existing diffusion-based Low-Light Image Enhancement (LLIE) approaches primarily train from scratch~\cite{yi2023diff,jiang2023low} (i.e., not leverage pre-trained large models). As a result, they remain restoration-based and may fail to produce satisfactory results in certain noisy regions, as illustrated in Fig.~\ref{fig:teaser} (a).
The potential of PTDB for LLIE with supervised setting is relatively underexplored, since significant information loss caused by dark environments further exacerbates fidelity concerns~\cite{xu2022snr, cai2023retinexformer} (examples are shown in Fig.~\ref{fig:teaser} (b)).
However, leveraging pre-trained diffusion models with strong generative capabilities for LLIE is a promising direction worth exploring, e.g., it can generate visually pleasing content in regions with severe noise. 
Several methods have been proposed to address fidelity in PTDB methods, with SR serving as a key example. Most approaches focus on designing more sophisticated conditioning networks~\cite{yu2024scaling,qu2024xpsr,chen2024cassr,xie2024addsr}. However, they often overlook the importance of optimizing the \textit{condition} itself for guiding the diffusion. 
If the provided condition lacks sufficient structural or semantic information for generation, PTDB models rely heavily on pre-trained knowledge, often producing content that deviates from the true details of the input image.

\begin{figure}[t]
    \centering
    \includegraphics[width=1\linewidth]{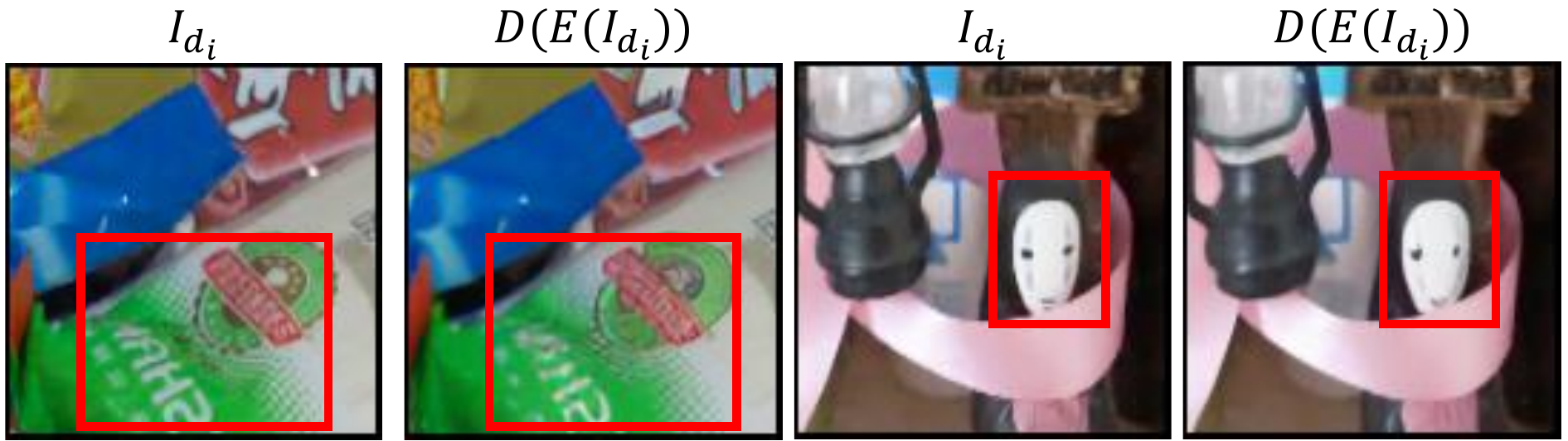}
    \caption{VAE reconstructions may contain distortions, meaning the conditional latent cannot fully capture the entire information of input image. Conditional latents should be refined.
    }
 \label{fig:teaser2}
\end{figure}

Obtaining the desired condition with enough information is challenging, as most pre-trained diffusion models operate in the latent space for efficiency.
To obtain the conditional latent, a VAE encoder is applied, which inevitably leads to spatial information loss, particularly in high-compression encoders (as shown in Fig.~\ref{fig:teaser2}). 
For instance, in Stable Diffusion~\cite{rombach2022high}, an input image of size $ H \times W \times 3 $ is compressed to a latent of $ \frac{H}{8} \times \frac{W}{8} \times 4 $, resulting in a nearly $48×$ compression ratio.  
Even though VAEs have undergone steady improvements (e.g., Flux~\cite{flux2024}) there is still no guarantee that the reconstruction error can be eliminated.

In this paper, we design novel plug-and-play conditional latent optimization strategy. First, we introduce a conditional latent refinement approach that transfers high-resolution information to the compressed latent representation. 
Second, we introduce a new bidirectional interaction mechanism between the refined latent and the noisy latent in the diffusion process. This mechanism enables effective information exchange, implicitly enhancing the formulated condition. 

Refining the conditional latent to address information loss from VAE compression is a highly ill-posed problem due to the more complex distribution in latent space compared to pixel space. To address this, we propose a two-stage strategy. First, we leverage a generative approach to establish an effective prior, expanding the solution space~\cite{ho2020denoising}. Then, this prior is used to predict the final refined latent representation, ensuring high-fidelity reconstruction.
In implementation, we begin with an information-lossless transformation to align the conditional input image with the compressed latent space. Next, a lightweight diffusion-based process generates the prior from the resized inputs, with a newly designed pyramidal target.
Finally, the predicted prior, combined with the resized data, guides the refinement under supervision from the ground-truth latent.
In this prediction process, a spatial-varying attention estimation mechanism is designed.
Compared to directly predicting the refined latent, our generative-prior-guided approach significantly improves accuracy in latent refinement.

Beyond latent refinement, we observe that previous approaches have overlooked the bidirectional feature interaction between the conditional latent and the noisy latent during the diffusion process. Existing methods primarily use the noisy latent at different time steps to constrain and progressively refine the conditional latent, incorporating time-dependent control~\cite{tsao2024holisdip,chen2024faithdiff,yu2024scaling}. However, we argue that the conditional latent, which retains fidelity that the noisy latent may lack, can also aid in refining the noisy latent throughout the process.
To address this, we propose a novel bidirectional feature interaction mechanism within the diffusion process, enabling mutual information exchange between the conditional latent and the noisy latent. This interaction enhances the effectiveness of both representations, leading to improved restoration (Fig.~\ref{fig:teaser}).

Extensive experiments are conducted on public datasets~\cite{yang2021sparse,chen2018learning,chen2019seeing} and across various PTDB methods. The results demonstrate that our proposed strategy follows a plug-and-play paradigm, effectively enhancing fidelity across different scenes while preserving the advantages of PTDB models (Tables~\ref{comparison1} and \ref{comparison1-unsupervised}).
In summary, our contribution is three-fold.
\begin{itemize}
    \item We propose a novel latent refinement strategy guided by a generative prior, effectively directing the degraded latent toward a high-quality representation.
    \item We identify the significance of bi-directional feature interaction between the conditional latent and noisy latent in restoration tasks and propose a corresponding method.
    \item Extensive experiments across various datasets and networks validate our proposed method.
\end{itemize}

\noindent\textbf{Clarification.}
In this paper, our method aims to \textbf{improve the fidelity of current PTDB strategies} to enable their practical use as a plug-and-play strategy. Two key clarifications are needed:
1) Reference-based metrics (e.g., PSNR, SSIM) of enhanced PTDB are not expected to surpass those of SOTA traditional restoration methods, as PTDB is fundamentally a generative approach rather than a restorative one. These methods serve different purposes and are not directly comparable in terms of fidelity, which is a widely acknowledged in the field~\cite{wu2024seesr,wang2024exploiting}. Therefore, outperforming traditional restoration models is not our goal.
2) Our primary focus is on enhancing conditional modeling in PTDB for restoration, which a relatively unexplored research topic. Efficiency optimization will be addressed in future work.

%% file: sec/2_related_work.tex
\section{Related Works}

\subsection{Restoration Models using Pre-trained Diffusion}

With the advancement of pre-trained diffusion models, particularly those designed for text-to-image generation, image restoration tasks have encountered new opportunities. Pre-trained models such as Stable Diffusion~\cite{rombach2022high} and Flux~\cite{flux2024} encompass a wealth of high-quality information that can significantly enhance restoration performance. Recent efforts in image super-resolution~\cite{wang2024exploiting,yang2024pixel,lin2024diffbir,wu2024one,sun2024coser,yu2024scaling,qu2024xpsr,sun2024pixel,chen2024faithdiff,luo2024photo,wang2024rap,dong2024tsd,qu2024xpsr} 
have largely concentrated on developing more effective control mechanisms.
StableSR~\cite{wang2024exploiting} represents the first attempt to implement the PTDB strategy, utilizing the ControlNet approach and optimizing fidelity in the VAE decoder. PASD~\cite{yang2024pixel} introduces a pixel-aware cross-attention module, enabling diffusion models to better capture local structures. DiffBIR~\cite{lin2024diffbir} also employs the ControlNet, but with the addition of a region-adaptive restoration guidance that modifies the denoising process during inference.
Despite these advancements, fidelity remains a significant challenge in PTDB strategies, with few efforts optimizing conditional latents.

\begin{figure*}[t]
	\begin{center}
		\includegraphics[width=.828\linewidth]{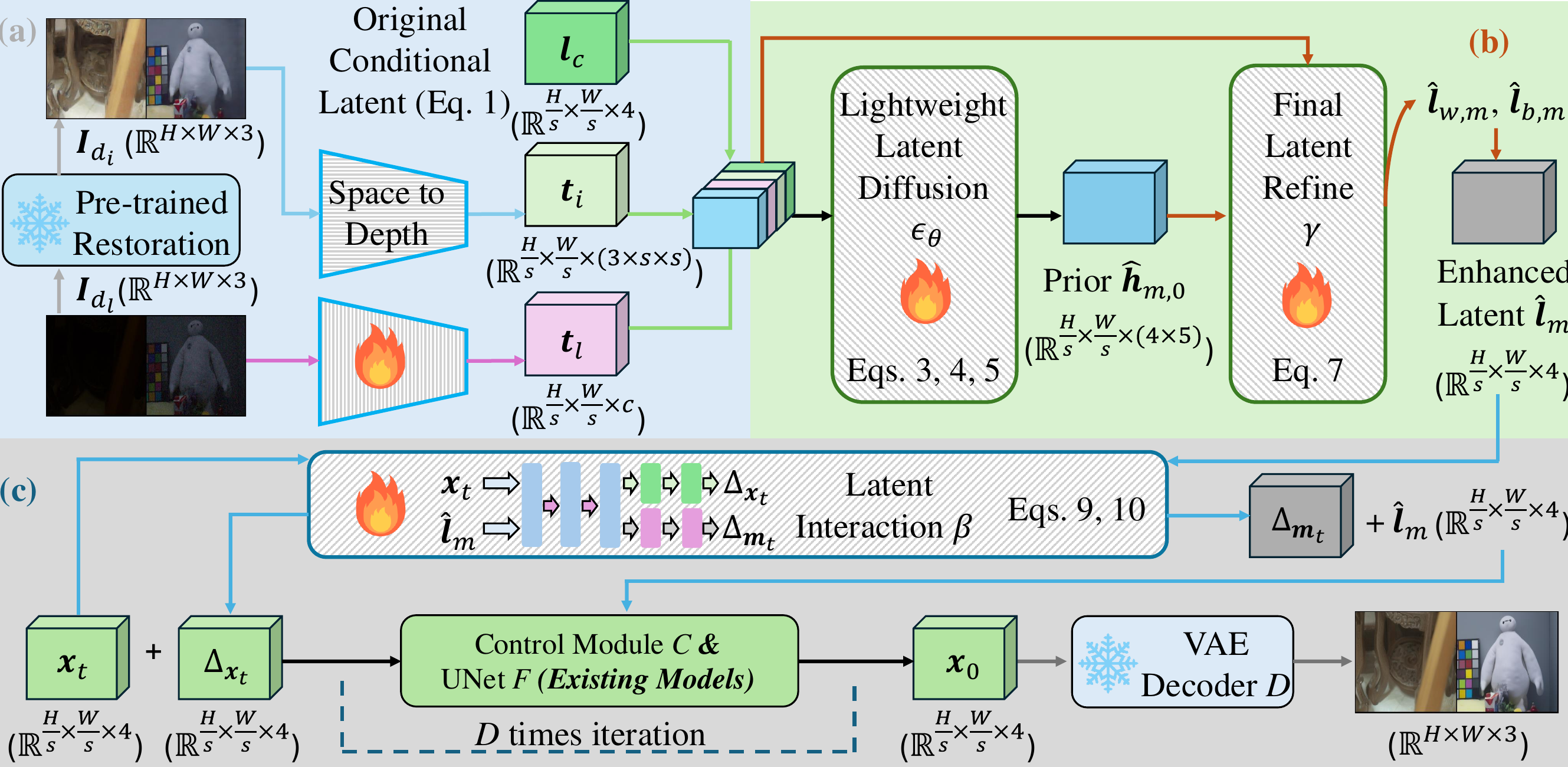}
	\end{center}
	\caption{
Illustration of our strategy for conditional latent refinement and interaction. (a) The input image and its enhanced version are used for refinement through information-lossless operations, such as space-to-depth and a visual spatial encoder. (b) These inputs, $ t_i $ and $ t_l $, feed into the latent refinement procedure, which includes the generative prior and final latent prediction, yielding $ \hat{h}_{m,0} $ and $ \hat{l}_m $. (c) The refined latent $ \hat{l}_m $ then interacts bidirectionally with the noisy latent $ x_t $ in the diffusion backbone $ F $, promoting residuals $ \Delta_{x_t} $ and $ \Delta_{m_t} $, 
thereby enhancing the performance of $ C $.
``$D$ times iteration" refers to $D$ diffusion steps.
	}
	\label{fig:framework}
\end{figure*}

\subsection{Low-light Image Enhancement via Diffusion}

Current approaches for low-light image enhancement primarily focus on improving network architectures~\cite{lin2023geometric,xu2025learnable,xu2025low,xu2023deep,xu2023low}.
Zamir et al.~\cite{zamir2022restormer} introduced Restormer, which captures long-range pixel interactions through attention mechanisms at the channel level. Xu et al.~\cite{xu2022snr} developed a network that combines convolutional and Transformer blocks in the latent space, integrated with an SNR map. 
Additionally, diffusion-based methods have been developed, with nearly all of them utilizing variants of diffusion models and training the model from scratch~\cite{yi2023diff,jiang2023low,hou2024global,wang2023exposurediffusion,jiang2024lightendiffusion,kang2024image,feng2024difflight,lv2024fourier,wu2024jores,chan2024anlightendiff}. 
E.g., Diff-L~\cite{jiang2023low} uses a wavelet-based conditional diffusion model that harnesses the generative power to produce results with satisfactory perceptual fidelity. 
Moreover, existing PTDB methods for low-light image enhancement are few and primarily designed for unsupervised settings, leveraging the generative capabilities of pre-trained diffusion models, such as QuadPrior~\cite{wang2024zero}, LLIEDiff~\cite{huang2025zero}, and LightenDiffusion~\cite{jiang2024lightendiffusion}. However, limited efforts have been devoted to exploring and advancing the use of large pre-trained diffusion models in supervised low-light image enhancement. This is a domain that demands high fidelity and holds substantial potential as diffusion models continue to evolve.

%% file: sec/3_method.tex
\section{Method}
\label{sec:method}

\subsection{PTDB Restoration Strategy}

\noindent\textbf{Normal strategy.} The PTDB strategy typically comprises two key components: the pre-trained diffusion network $F$ (e.g., UNet~\cite{rombach2022high}, DiT~\cite{peebles2023scalable}) and the condition used for control (e.g., input low-quality data or initialized enhanced data). Let $\boldsymbol{I}_{d_l}$ represent the original low-light input, $\boldsymbol{I}_{d_i}$ is the initialized enhanced data, and $\boldsymbol{I}_n$ denotes the corresponding high-quality data.
The restoration procedure can be denoted as 
\begin{equation}
    \small
\hat{\boldsymbol{I}}_h=D(F(\boldsymbol{\mathcal{N}}, C(\boldsymbol{l}_c))), \; \boldsymbol{l}_c= E(\boldsymbol{I}_{d_l}) \; \mathrm{or} \; \boldsymbol{l}_c= E(\boldsymbol{I}_{d_i}),
\end{equation}
where $\boldsymbol{\mathcal{N}}$ denotes the sampled Gaussian noise, $E$ is the VAE encoder, $D$ is the VAE decoder, $C$ is the conditional module that controls the generation process of $F$ using the conditional latent $\boldsymbol{l}_c$, and $\hat{\boldsymbol{I}}_h$ the predicted output. The employment of the VAE encoder and decoder is primarily motivated by efficiency requirements and ease of training. The implementation of the conditional mechanism $C$ is mainly inspired by the ControlNet~\cite{zhang2023adding}, either inserting the conditional latent directly or concatenating it with the noisy latent~\cite{chen2024faithdiff}.

One critical issue exists in previous strategies (although they are not validated in low-light enhancement task): insufficient fidelity. For example, a person's identity or the color/style of a scene may be altered. This issue stems from the limitations of the control module $C$ and the modeling of the conditional latent $\boldsymbol{l}_c$. Some previous works have focused on enhancing the mechanism of $C$, e.g., incorporating Lora layers with ControlNet~\cite{wang2024hero}, adopting dynamic mechanisms for ControlNet at different time steps~\cite{yang2024pixel}, and modifying its structure~\cite{qu2024xpsr,chen2024cassr}. However, modeling conditional latent $\boldsymbol{l}_c$ is also vital.

\noindent\textbf{The motivation for the conditional latent refinement.}
The conditional latent representation is obtained by applying a VAE encoder to the input low-quality image or an initially enhanced image, which involves a compression process. Given that the shape of $ \boldsymbol{I}_{d_l}$ and $\boldsymbol{I}_{d_i} $ is $ H \times W \times 3 $, the resulting conditional latent representation $ \boldsymbol{l}_c $ will have a shape of $ \frac{H}{s} \times \frac{W}{s} \times 4 $~\cite{rombach2022high}, where $ s $ denotes the compression ratio.  
Although the VAE decoder can approximately reconstruct the original image $ \boldsymbol{I}_{d_{l,i}} $ from $ \boldsymbol{l}_c $, the process inevitably introduces reconstruction distortions or errors. Furthermore, some spatial information is ineluctably lost during compression, meaning that $ \boldsymbol{l}_c $ cannot fully preserve the information in $ \boldsymbol{I}_{d_{l,i}} $ for effectively controlling the behavior of $ C $. Consequently, this impacts the fidelity of the generated images, leading to more severe fidelity issues in final outputs after decoding. This issue is particularly severe in low-light image enhancement tasks, where low-light images are noisy and significant information is damaged.

Therefore, we propose a method to compensate for the bereft information in the conditional latent representation by incorporating missing details from $ \boldsymbol{I}_{d_i} $ and $ \boldsymbol{I}_{d_l} $ into $ \boldsymbol{l}_c $. In other words, our goal is to refine the conditional latent. Furthermore, we observe that the closer $ \boldsymbol{l}_c $ is to the latent representation of the ground truth, the better the generated results. Thus, we set the refinement target to the ground truth’s latent representation, defined as $ \boldsymbol{l}_m = E(\boldsymbol{I}_n) $.  The strategy can be viewed in Fig.~\ref{fig:framework}.

\begin{figure}[t]
    \centering
\includegraphics[width=1\linewidth]{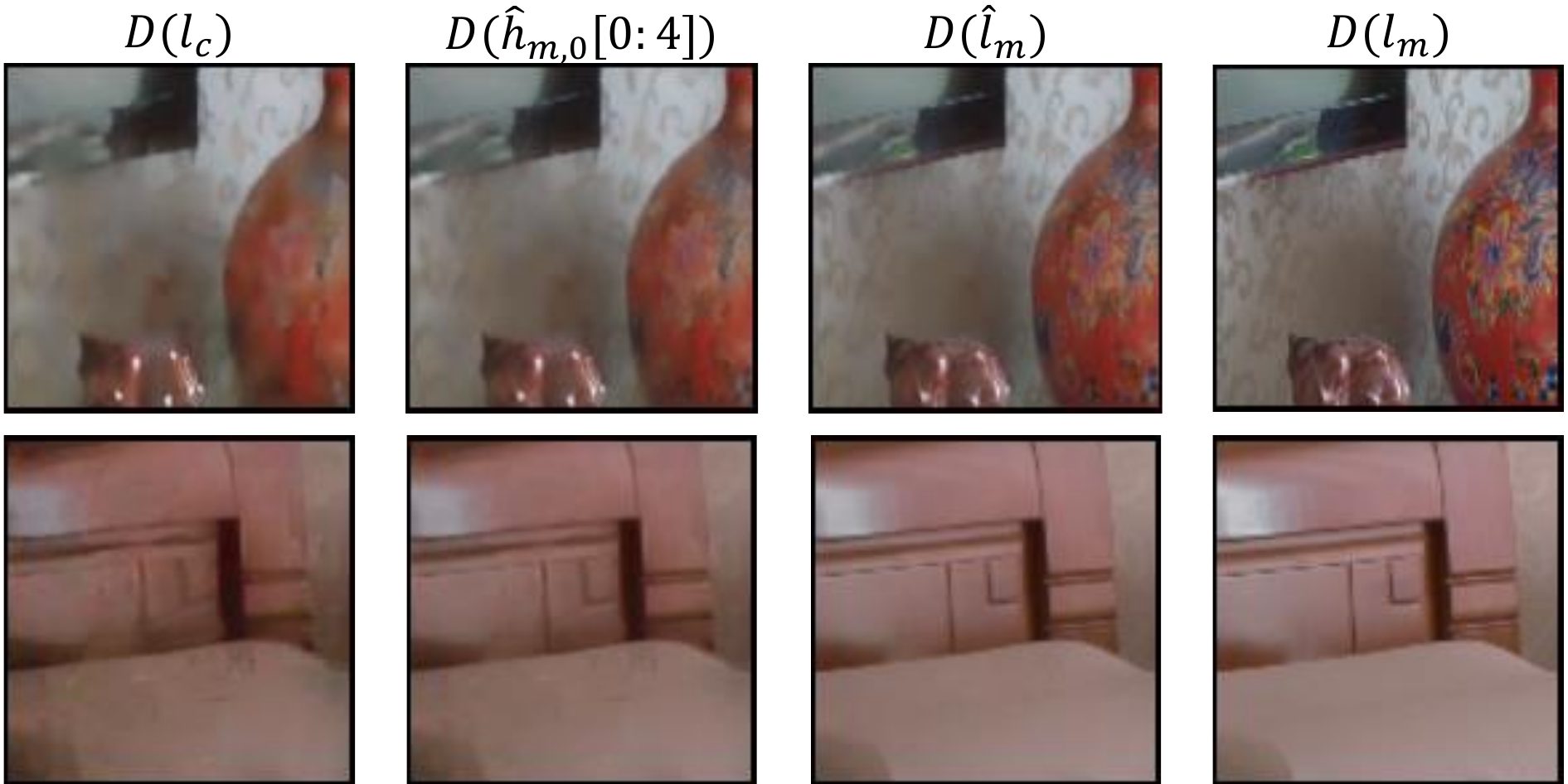}
 \caption{Visual samples to show the effects of our latent refinement strategy. 
 We decode the generative prior $\hat{\boldsymbol{h}}_{m,0}$ (at its highest scale, i.e., the first four channels), the refined latent $\hat{\boldsymbol{l}}_m$, and the ground truth latent. The results show that the prior indicates the correct improvement direction, and the refined latent is closer to the ground truth.
 }
    \label{fig:visual}
\end{figure}

\subsection{The Latent Refinement Strategy}

Refining the latent representation is a challenging problem. Unlike pixel-level refinement, the latent space has an extremely complex distribution due to being trained with both a reconstruction loss and a KL divergence constraint~\cite{rombach2022high}, making the refinement process highly ill-posed.  
To address this challenge, we propose a two-stage strategy. First, we leverage the generative capability of the diffusion model~\cite{ho2020denoising} to obtain a suitable prior, which can approximate the distribution of $ \boldsymbol{l}_m=E(\boldsymbol{I}_n) $ while allowing for diverse solutions. Then, this prior, combined with the information from $ \boldsymbol{I}_{d_i} $ and $ \boldsymbol{I}_{d_l} $, is used to regress a more accurate conditional latent representation. This refined latent, which closely approximates $ \boldsymbol{l}_m = E(\boldsymbol{I}_n) $ with high fidelity, enhances downstream restoration performance. In the following sections, we introduce these two stages in detail.

\noindent\textbf{The suitable input to formulate priors.}
First, it is crucial to utilize the full content of the input image to compensate for the spatial information lost during VAE encoder compression (Fig.~\ref{fig:teaser2}). We think the information can be sourced from both the original low-light image and the initially enhanced image, i.e., $ E(\boldsymbol{I}_{d_i}) $ and $ E(\boldsymbol{I}_{d_l}) $.
Instead of relying on a compression-based approach, we adopt another strategy to construct a reliable input for obtaining the prior. Specifically, for the original low-light image, we introduce a visual spatial encoder that transforms $ \boldsymbol{I}_{d_l} $ into a tensor of size $ \boldsymbol{t}_l \in \mathbb{R}^{\frac{H}{s} \times \frac{W}{s} \times c }$, where $c$ is large enough.
The purpose of this visual encoder is to filter out degradation artifacts in $ \boldsymbol{I}_{d_l} $.  
For the initially enhanced image $ \boldsymbol{I}_{d_i} $, we directly apply a pixel shuffle operation (i.e., space to depth)~\cite{shi2016real}, as the degradation factors have already been removed at the pixel level. The resulting tensor has a shape of $ \boldsymbol{t}_i \in \mathbb{R}^{\frac{H}{s} \times \frac{W}{s} \times c'} $, where $ c' = 3 \times s \times s $.  

After obtaining these inputs, which contain valuable information to compensate for the loss in the encoder, we employ a diffusion model to generate a prior. The diffusion process ensures that the prior generation aligns well with the characteristics of the pre-trained diffusion model $ F $, which also involves a Gaussian sampling strategy in the latent space. By leveraging this prior, we can guide the model toward obtaining a suitable conditional latent representation, enabling the diffusion model to better fulfill our fidelity requirements.

\begin{table}[t]
	\centering
	\resizebox{1\linewidth}{!}{
		\begin{tabular}{|c|c|c|c|c|c|c|}
			\hline
			Layer Type 		 & Norm & Activation & Kernel & Stride & Padding & Output Size \\ 
			\hline \hline
			Input Feature       	 & -    & -       & -      & -   &-   & $H \times W \times 3$ \\ \hline
			Convolution 	 & - & LeakyReLU    & 4      & 2   &1   & $H/2 \times W/2 \times 64$ \\ \hline
			SCUNet Block 	 & - & LeakyReLU    & -      & -   &-  & $H/2 \times W/2 \times 64$ \\ \hline
			Convolution 	 & - & LeakyReLU    & 4      & 2   &1   & $H/4 \times W/4 \times 128$ \\ \hline
			SCUNet Block 	 & - & LeakyReLU    & -      & -   &-  & $H/4 \times W/4 \times 128$ \\ \hline
			Convolution 	 & - & LeakyReLU    & 4      & 2   &1   & $H/8 \times W/8 \times 256$ \\ \hline
			SCUNet Block 	 & - & LeakyReLU    & -      & -   &-  & $H/8 \times W/8 \times 256$ \\ \hline
			Convolution 	 & - & LeakyReLU    & 3      & 1   &1   & $H/8 \times W/8 \times 256$ \\ \hline
			Convolution 	 & - & -    & 1      & 1   &0   & $H/8 \times W/8 \times 64$ \\ 
			\hline
	\end{tabular}}
        	\caption{Architecture of the visual encoder $E_v$ to obtain $\boldsymbol{t}_l$.} 
	\label{tab:visual-encoder}
\end{table}

\begin{table}[t]
	\centering
	\resizebox{1\linewidth}{!}{
		\begin{tabular}{|c|c|c|c|c|c|c|}
			\hline
			Layer Type 		 & Norm & Activation & Kernel & Stride & Padding & Output Size \\ 
			\hline \hline
			Input Feature       	 & -    & -       & -      & -   &-   & $H \times W \times C$ \\ \hline
			Convolution 	 & - & -    & 3      & 1   &1   & $H \times W \times 256$ \\ \hline
			SCUNet Block 	 & - & LeakyReLU    & -      & -   &-  & $H \times W \times 256$ \\ \hline
			Convolution 	 & - & LeakyReLU    & 3      & 1   &1   & $H \times W \times 256$ \\ \hline
			SCUNet Block 	 & - & LeakyReLU    & -      & -   &-  & $H \times W \times 256$ \\ \hline
			Convolution 	 & - & LeakyReLU    & 3      & 1   &1   & $H \times W \times 256$ \\ \hline
			SCUNet Block 	 & - & LeakyReLU    & -      & -   &-  & $H \times W \times 256$ \\ \hline
			Convolution 	 & - & LeakyReLU    & 3      & 1   &1   & $H \times W \times 256$ \\ \hline
			Convolution 	 & - & -    & 3      & 1   &1   & $H \times W \times 20$ \\ 
			\hline
	\end{tabular}}
        \caption{Architecture of the denoising network in the latent diffusion model $\epsilon_\theta$.} 
	\label{tab:denoise-diffusion}
\end{table}
\begin{table}[t]
		\resizebox{1\linewidth}{!}{
		\begin{tabular}{|c|c|c|c|c|c|c|}
			\hline
			Layer Type 		 & Norm & Activation & Kernel & Stride & Padding & Output Size \\ 
			\hline \hline
			Input Feature       	 & -    & -       & -      & -   &-   & $H \times W \times C$ \\ \hline
			Convolution 	 & - & -    & 3      & 1   &1   & $H \times W \times 256$ \\ \hline
			SCUNet Block 	 & - & LeakyReLU    & -      & -   &-  & $H \times W \times 256$ \\ \hline
			Convolution 	 & - & LeakyReLU    & 3      & 1   &1   & $H \times W \times 256$ \\ \hline
			Convolution 	 & - & -    & 1      & 1   &0  & $H \times W \times 256$ \\ 
			\hline
	\end{tabular}}
        	\caption{Architecture of the conditional network in the latent diffusion model $\epsilon_\theta$.} 
	\label{tab:condition-diffusion}
\end{table}

\noindent\textbf{The modeling of priors.}
Unlike traditional solutions that directly generate the target $\boldsymbol{l}_m$, we propose synthesizing the target latent representation at multiple scales, forming a pyramid structure. Notably, the lower the scale, the less ill-posed the problem becomes. Moreover, in this framework, accurate predictions at lower scales may effectively help generation at higher scales. Given $\boldsymbol{l}_m=E(\boldsymbol{I}_n)$, the generation target, serving as the ground truth, can be expressed as follows
\begin{equation}
\begin{aligned}
\boldsymbol{h}_m= \boldsymbol{l}_m &\oplus \uparrow_2(\downarrow_2(\boldsymbol{l}_m)) \oplus \uparrow_4(\downarrow_4(\boldsymbol{l}_m)) \\
        &\oplus \uparrow_8(\downarrow_8(\boldsymbol{l}_m)) \oplus \uparrow_{16}(\downarrow_{16}(\boldsymbol{l}_m)),
\end{aligned}
\label{eq:pyramid}
\end{equation}
where $\oplus$ represents the channel concatenation operation, while $\uparrow_b$ and $\downarrow_b$ denote bilinear upsampling and downsampling operations with a scale factor of $b$, respectively.

To perform the diffusion process, we first use the clean $\boldsymbol{h}_m$ to sample $\boldsymbol{h}_{m,T}$, as
\begin{equation}
    q(\boldsymbol{h}_{m,T}|\boldsymbol{h}_m) = \boldsymbol{\mathcal{N}}(\boldsymbol{h}_{m,T}; \sqrt{\bar{\alpha}_T} \boldsymbol{h}_m, (1-\bar{\alpha}_T)\boldsymbol{I}),
\end{equation}
where $T$ is the total number of diffusion iteration, $\bar{\alpha}_T$ and $\alpha_T$ are the parameters in DDPM~\cite{ho2020denoising}, $\boldsymbol{\mathcal{N}}$ is the Gaussian distribution.
Then, in the reverse process, we start from the $T$-th time step and perform all denoising iterations to obtain the diffusion output, following the strategy of DiffIR~\cite{xia2023diffir}, as
\begin{equation}
    \hat{\boldsymbol{h}}_{m,t-1}=(1/\sqrt{\alpha_t})(\hat{\boldsymbol{h}}_{m,t}-\boldsymbol{\epsilon} ((1-\alpha_t)/{\sqrt{1-\bar{\alpha}_t}})), 
\end{equation}
where $\hat{\boldsymbol{l}}_{m,0}$ is the obtained prior that is expected to be close to $\boldsymbol{l}_m=E(\boldsymbol{I}_n)$.
$\epsilon$ is predicted by the network with the constructed conditions, as
\begin{equation}
    \boldsymbol{\epsilon}=\epsilon_\theta(\hat{\boldsymbol{h}}_{m,t},t, \boldsymbol{t}_i, \boldsymbol{t}_l, \boldsymbol{l}_c),
    \label{eq:eps}
\end{equation}
where $\boldsymbol{t}_i, \boldsymbol{t}_l, \boldsymbol{l}_c$ are all conditions.
The training and inference procedures follow the same sampling path.
To supervise the learning of this network, we apply a MSE loss function, as
\begin{equation}
\begin{aligned}
        &\mathcal{L}_\text{g} = \mathbb{E}(\Vert \hat{\boldsymbol{h}}_{m,0}-\boldsymbol{h}_m \Vert).
\end{aligned}
\label{eq:refine1}
\end{equation}

\noindent\textbf{Predict the refinement with the prior.}
Although the obtained prior is close to the target, it still contains some variances, which could introduce errors in controlling the generation, as shown in Fig.~\ref{fig:visual}. Fortunately, once the prior is obtained, we can significantly weaken the high ill-posedness issue by using the prior as a condition for lightweight regression.
Specifically, we adopt an attention-aware prediction approach, assuming that certain regions of the conditional latent $ \boldsymbol{l}_c $ are already satisfied. Therefore, optimization is focused only on the regions that require improvement.
Thus, we set a another network as $\gamma$ to predict 
\begin{equation}
        \hat{\boldsymbol{l}}_{w,m}, \hat{\boldsymbol{l}}_{b,m} = \gamma(\hat{\boldsymbol{h}}_{m,0}, \boldsymbol{t}_i, \boldsymbol{t}_l, \boldsymbol{l}_c),\,
\hat{\boldsymbol{l}}_{m}=\boldsymbol{l}_c+\hat{\boldsymbol{l}}_{w,m}\odot \hat{\boldsymbol{l}}_{b,m},
     \label{eq:lm}
\end{equation}
where $\hat{\boldsymbol{l}}_{w,m}$ is the weighting map with the value belonging to $[0,1]$, and $\odot$ is the Hadamard product.
A MSE loss is also applied, as
\begin{equation}
\begin{aligned}
        &\mathcal{L}_\text{r} = \mathbb{E}(\Vert \hat{\boldsymbol{l}}_{m}-\boldsymbol{l}_m \Vert_2).
\end{aligned}
\label{eq:refine}
\end{equation}

\begin{table}[t]
	\centering
	\resizebox{1\linewidth}{!}{
		\begin{tabular}{|c|c|c|c|c|c|c|}
			\hline
			Layer Type 		 & Norm & Activation & Kernel & Stride & Padding & Output Size \\ 
			\hline \hline
			Input Feature       	 & -    & -       & -      & -   &-   & $H \times W \times C$ \\ \hline
			Convolution 	 & - & -    & 3      & 1   &1   & $H \times W \times 512$ \\ \hline
			SCUNet Block 	 & - & LeakyReLU    & -      & -   &-  & $H \times W \times 512$ \\ \hline
			Convolution 	 & - & LeakyReLU    & 3      & 1   &1   & $H \times W \times 256$ \\ \hline
			Convolution 	 & - & -    & 1      & 1   &0  & $H \times W \times 256$ \\ 
			\hline
	\end{tabular}}
        	\caption{Architecture of the shared part in final latent predictor $\gamma$.} 
	\label{tab:prediction}
\end{table}

\begin{table}[t]
	\centering
	\resizebox{1\linewidth}{!}{
		\begin{tabular}{|c|c|c|c|c|c|c|}
			\hline
			Layer Type 		 & Norm & Activation & Kernel & Stride & Padding & Output Size \\ 
			\hline \hline
			Input Feature       	 & -    & -       & -      & -   &-   & $H \times W \times 256$ \\ \hline
			Convolution 	 & - & -    & 3      & 1   &1   & $H \times W \times 128$ \\ \hline
			SCUNet Block 	 & - & LeakyReLU    & -      & -   &-  & $H \times W \times 128$ \\ \hline
			Convolution 	 & - & LeakyReLU    & 3      & 1   &1   & $H \times W \times 64$ \\ \hline
			Convolution 	 & - & -    & 1      & 1   &0  & $H \times W \times 2$ \\ 
			\hline
	\end{tabular}}
        	\caption{Architecture of the individual output head for each channel in the final latent predictor $\gamma$.} 
	\label{tab:prediction2}

\end{table}

\begin{table}[t]
	\centering
	\resizebox{1\linewidth}{!}{
		\begin{tabular}{|c|c|c|c|c|c|c|}
			\hline
			Layer Type 		 & Norm & Activation & Kernel & Stride & Padding & Output Size \\ 
			\hline \hline
			Input Feature       	 & -    & -       & -      & -   &-   & $H \times W \times 8$ \\ \hline
			Convolution 	 & - & -    & 3      & 1   &1   & $H \times W \times 256$ \\ \hline
			SCUNet Block 	 & - & LeakyReLU    & -      & -   &-  & $H \times W \times 256$ \\ \hline
			Convolution 	 & - & LeakyReLU    & 3      & 1   &1   & $H \times W \times 256$ \\ \hline
			Convolution 	 & - & -    & 1      & 1   &0  & $H \times W \times 128$ \\ 
			\hline
	\end{tabular}}
        	\caption{Architecture of the shared part in the latent interaction module $\beta$.} 
	\label{tab:trans}
\end{table}

\begin{table}[t]
		\resizebox{1\linewidth}{!}{
		\begin{tabular}{|c|c|c|c|c|c|c|}
			\hline
			Layer Type 		 & Norm & Activation & Kernel & Stride & Padding & Output Size \\ 
			\hline \hline
			Input Feature       	 & -    & -       & -      & -   &-   & $H \times W \times 128$ \\ \hline
			Convolution 	 & - & -    & 3      & 1   &1   & $H \times W \times 512$ \\ \hline
			Convolution 	 & - & LeakyReLU    & 3      & 1   &1   & $H \times W \times 64$ \\ \hline
			Convolution 	 & - & -    & 1      & 1   &0  & $H \times W \times 4$ \\ 
			\hline
	\end{tabular}}
	\caption{Architecture of output head in latent interaction module $\beta$, predicting either $\Delta_{\boldsymbol{x}_t}$ or $\Delta_{\boldsymbol{m}_t}$.} 
	\label{tab:trans2}
\end{table}

\subsection{The Latent Interaction Strategy}
\label{sec:inter}

After obtaining a refined conditional latent representation, we emphasize the significance of the mutual interaction between the conditional latent and the noisy latent in the diffusion process of $ F $. Previous approaches have typically used a constant conditional latent throughout the diffusion model at different steps, without considering the distinct effects of noisy latents at varying stages. For instance, during the early stages of the diffusion process, the noisy latent is predominantly influenced by noise, necessitating a stronger conditional latent to provide meaningful information. In contrast, as the process progresses toward $ t = 0 $, the noisy latent primarily contains generated content with less noise, allowing it to contribute to a more controllable input. This underscores the need for bidirectional interaction.

Let the noisy latent in the diffusion process of $ F $ be denoted as $ \boldsymbol{x}_t $, and the conditional latent be represented by the predicted $ \hat{\boldsymbol{l}}_m $.
The bidirectional interaction is expressed as
\begin{equation}
    \Delta_{\boldsymbol{x}_t}, \Delta_{\boldsymbol{m}_t} = \beta(\boldsymbol{x}_t, \hat{\boldsymbol{l}}_m),
    \label{eq:inter1}
\end{equation}
where $\beta$ is the latent interaction module,
$\Delta_{\boldsymbol{x}_t}$ and $\Delta_{\boldsymbol{m}_t}$ represent the predicted residuals for the noisy and conditional latents, respectively. In this context, the denoising process in the diffusion network $F$ can be reformulated as follows
\begin{equation}
    F(\boldsymbol{x}_t, C(\boldsymbol{l}_m)) \xrightarrow{} F(\boldsymbol{x}_t+\Delta_{\boldsymbol{x}_t}, C(\boldsymbol{l}_m+\Delta_{\boldsymbol{m}_t})).
    \label{eq:inter2}
\end{equation}
The residual can be learned autonomously without the need for explicit supervision. We observe that the interaction mechanism enhances the control ability.

Note that our proposed latent refinement and interaction strategy is plug-and-play, allowing it to be seamlessly integrated with existing pre-trained diffusion-based networks.

\subsection{Training Loss}

The training loss of our method consists of two components. The first component is the original diffusion loss, such as the loss function used for ControlNet~\cite{zhang2023adding}, which we denote as $\mathcal{L}_{\text{diff}}$. The second component is the supervision loss for latent refinement. Besides the loss functions in Eqs.~\eqref{eq:refine1} and \eqref{eq:refine} that define the refinement in the latent space, we also introduce a constraint at the pixel level. This further encourages the refinement to preserve sufficient original information from the input images.
The loss is
\begin{equation}
\begin{aligned}
        &\mathcal{L}_{\text{gp}} = \mathbb{E}(\Vert D(\hat{\boldsymbol{h}}_{m,0})-D(\boldsymbol{h}_m) \Vert),\\
        &\mathcal{L}_{\text{rp}} = \mathbb{E}(\Vert D(\hat{\boldsymbol{l}}_{m})-D(\boldsymbol{l}_m) \Vert),
\end{aligned}
\label{eq:refine2}
\end{equation}
where $D$ is the VAE decoder.
In summary, the overall loss function can be written as
\begin{equation}
    \mathcal{L}_\text{all}=\mathcal{L}_\text{diff}+\lambda_1 (\mathcal{L}_\text{g}+\mathcal{L}_\text{r})+\lambda_2 (\mathcal{L}_\text{gp} +\mathcal{L}_\text{rp}),
\end{equation}
where $\lambda_1$ and $\lambda_2$ are loss weights, and they are set as 1 in this paper (their optimal values can be efficiently determined via a grid search over a limited range of candidate settings in practice).
Our code and models will be made publicly available upon publication, along with detailed implementation information.

%% file: sec/4_experiment.tex
\section{Experiments}
\label{sec:exp}

\begin{figure}[t]
	\centering

	\begin{subfigure}[c]{0.24\linewidth}
		\centering
		\includegraphics[width=0.88in,height=0.88in]{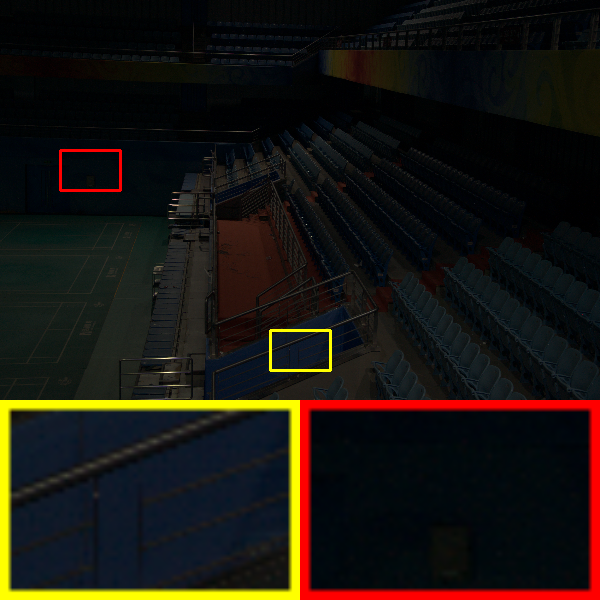}
		\vspace{-1.5em}
		\caption*{LOL-real}
	\end{subfigure}
	\begin{subfigure}[c]{0.24\linewidth}
		\centering
		\includegraphics[width=0.88in,height=0.88in]{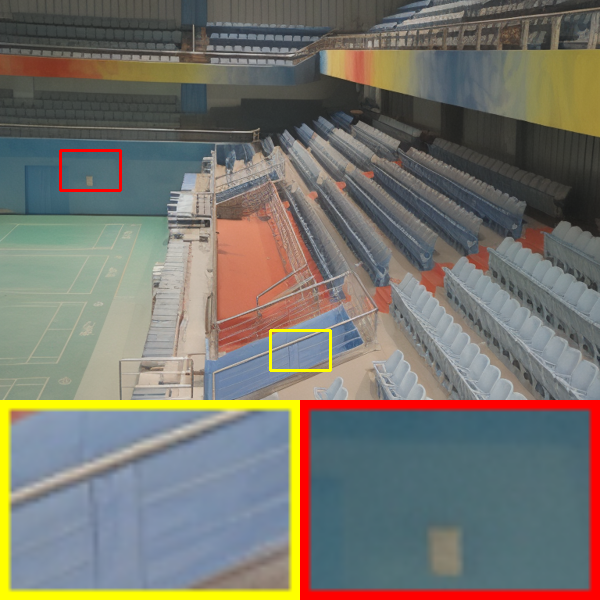}
		\vspace{-1.5em}
		\caption*{DiffBIR}
	\end{subfigure}
	\begin{subfigure}[c]{0.24\linewidth}
		\centering
		\includegraphics[width=0.88in,height=0.88in]{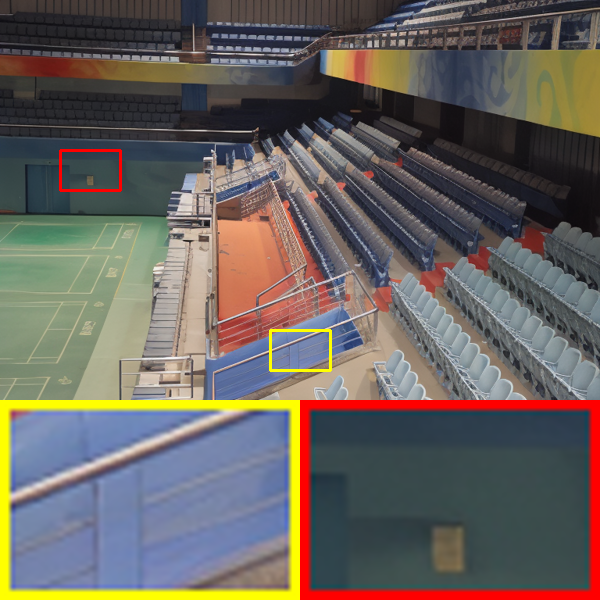}
		\vspace{-1.5em}
		\caption*{+Ours}
	\end{subfigure}
	\begin{subfigure}[c]{0.24\linewidth}
		\centering
		\includegraphics[width=0.88in,height=0.88in]{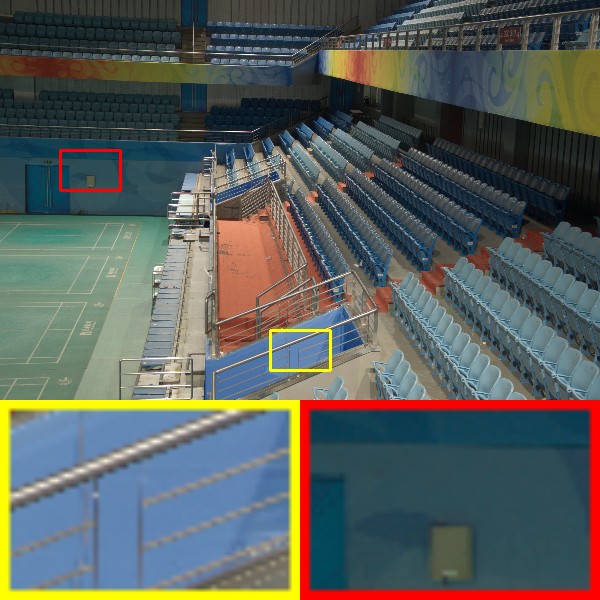}
		\vspace{-1.5em}
		\caption*{GT}
	\end{subfigure}

\begin{subfigure}[c]{0.24\linewidth}
		\centering
		\includegraphics[width=0.88in,height=0.88in]{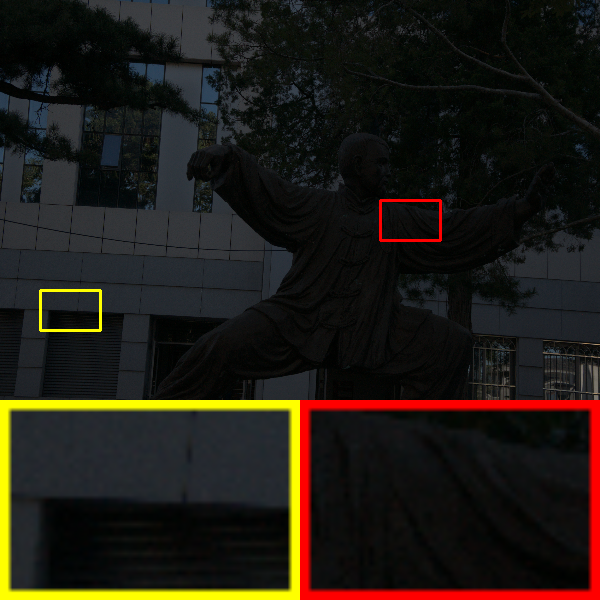}
		\vspace{-1.5em}
		\caption*{LOL-real}
	\end{subfigure}
	\begin{subfigure}[c]{0.24\linewidth}
		\centering
		\includegraphics[width=0.88in,height=0.88in]{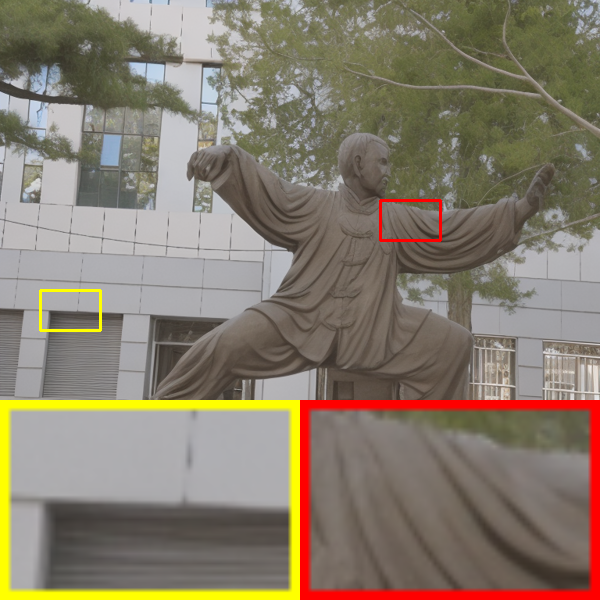}
		\vspace{-1.5em}
		\caption*{StableSR}
	\end{subfigure}
	\begin{subfigure}[c]{0.24\linewidth}
		\centering
		\includegraphics[width=0.88in,height=0.88in]{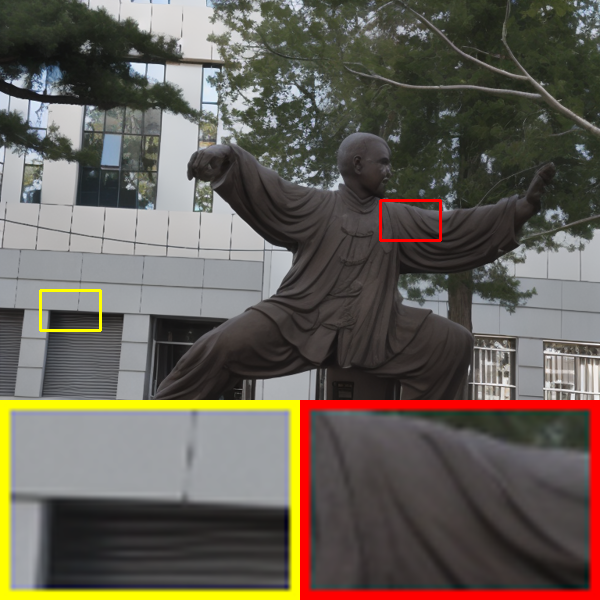}
		\vspace{-1.5em}
		\caption*{+Ours}
	\end{subfigure}
	\begin{subfigure}[c]{0.24\linewidth}
		\centering
		\includegraphics[width=0.88in,height=0.88in]{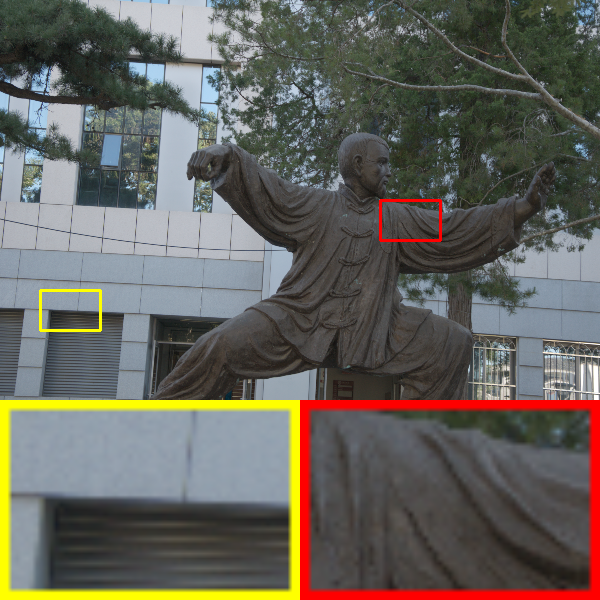}
		\vspace{-1.5em}
		\caption*{GT}
	\end{subfigure}

\begin{subfigure}[c]{0.24\linewidth}
		\centering
		\includegraphics[width=0.88in,height=0.88in]{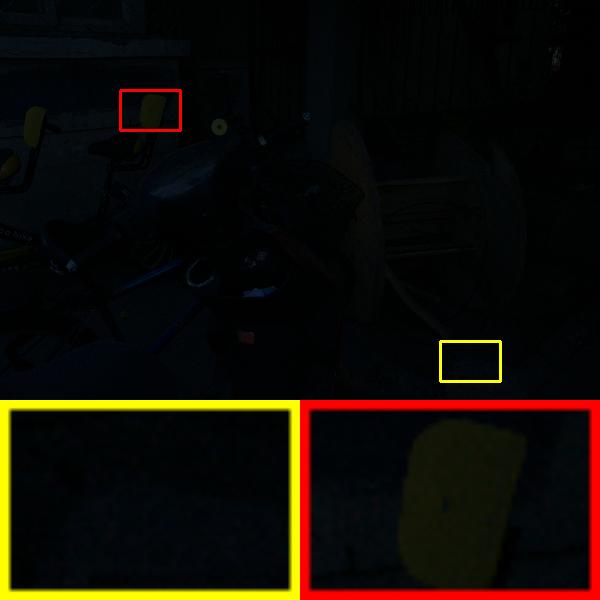}
		\vspace{-1.5em}
		\caption*{LOL-real}
	\end{subfigure}
	\begin{subfigure}[c]{0.24\linewidth}
		\centering
		\includegraphics[width=0.88in,height=0.88in]{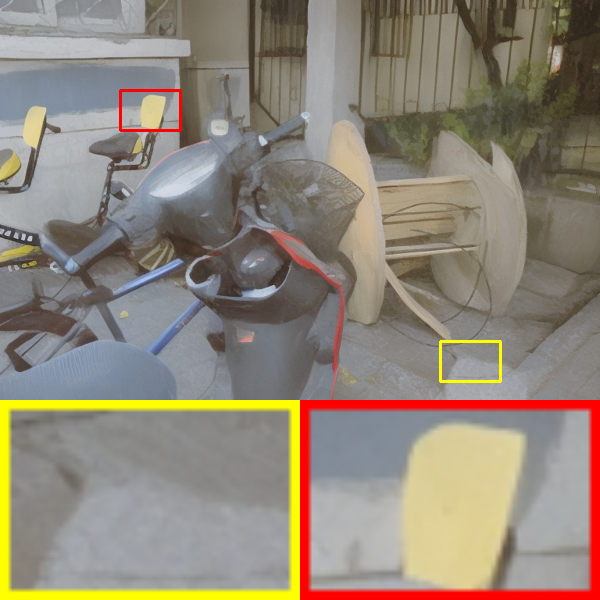}
		\vspace{-1.5em}
		\caption*{PASD}
	\end{subfigure}
	\begin{subfigure}[c]{0.24\linewidth}
		\centering
		\includegraphics[width=0.88in,height=0.88in]{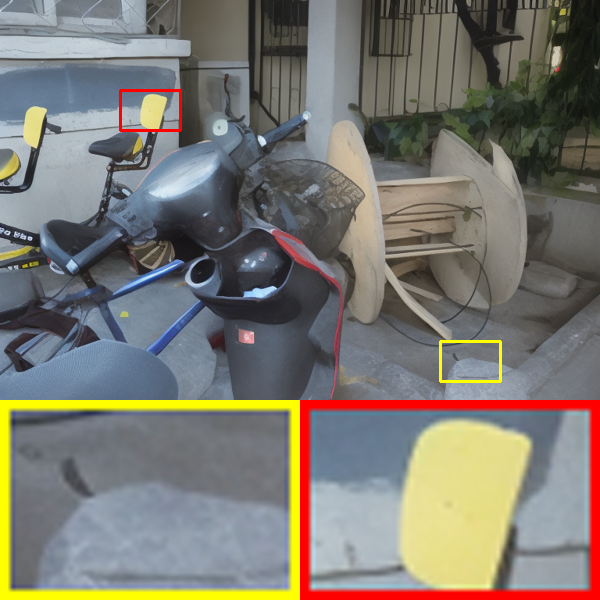}
		\vspace{-1.5em}
		\caption*{+Ours}
	\end{subfigure}
	\begin{subfigure}[c]{0.24\linewidth}
		\centering
		\includegraphics[width=0.88in,height=0.88in]{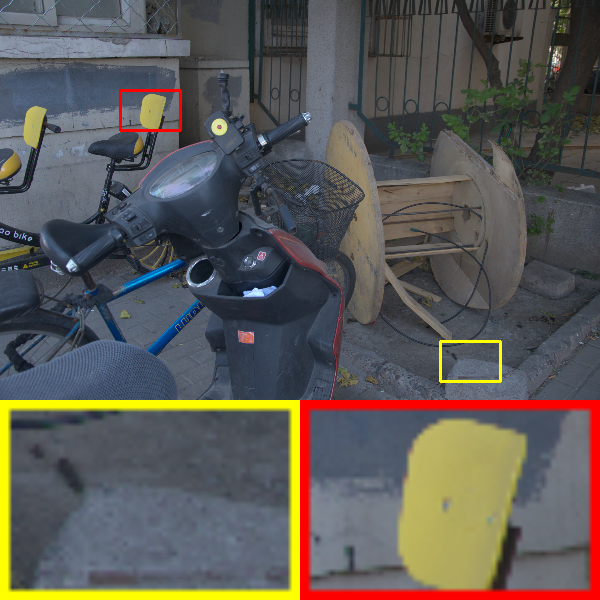}
		\vspace{-1.5em}
		\caption*{GT}
	\end{subfigure}

	\caption{Visual comparisons on different datasets with various network structures on LOL-real. 
    }
	\label{fig:cmp}
\end{figure}

\begin{figure}[t]
	\centering
	\begin{subfigure}[c]{0.24\linewidth}
		\centering
		\includegraphics[width=0.88in,height=0.88in]{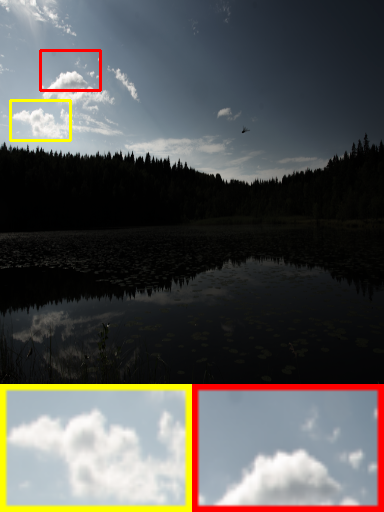}
		\vspace{-1.5em}
		\caption*{LOL-syn.}
	\end{subfigure}
	\begin{subfigure}[c]{0.24\linewidth}
		\centering
		\includegraphics[width=0.88in,height=0.88in]{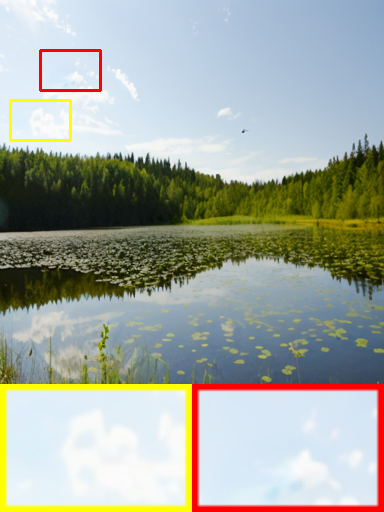}
		\vspace{-1.5em}
		\caption*{DiffBIR}
	\end{subfigure}
	\begin{subfigure}[c]{0.24\linewidth}
		\centering
		\includegraphics[width=0.88in,height=0.88in]{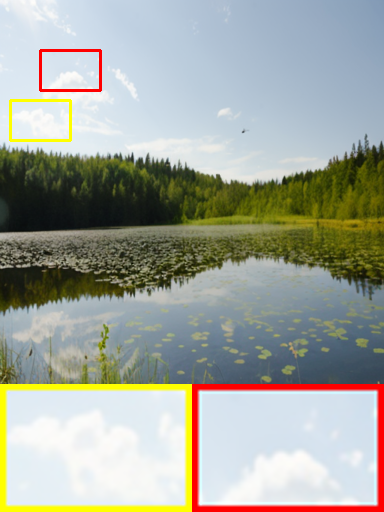}
		\vspace{-1.5em}
		\caption*{+Ours}
	\end{subfigure}
	\begin{subfigure}[c]{0.24\linewidth}
		\centering
		\includegraphics[width=0.88in,height=0.88in]{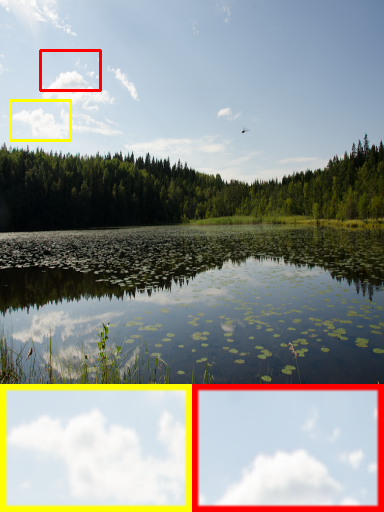}
		\vspace{-1.5em}
		\caption*{GT}
	\end{subfigure}

	\begin{subfigure}[c]{0.24\linewidth}
		\centering
		\includegraphics[width=0.88in,height=0.88in]{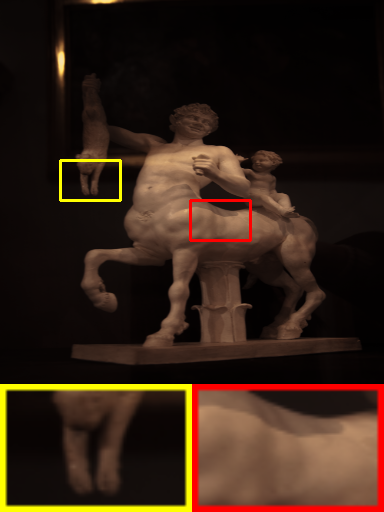}
		\vspace{-1.5em}
		\caption*{LOL-syn.}
	\end{subfigure}
	\begin{subfigure}[c]{0.24\linewidth}
		\centering
		\includegraphics[width=0.88in,height=0.88in]{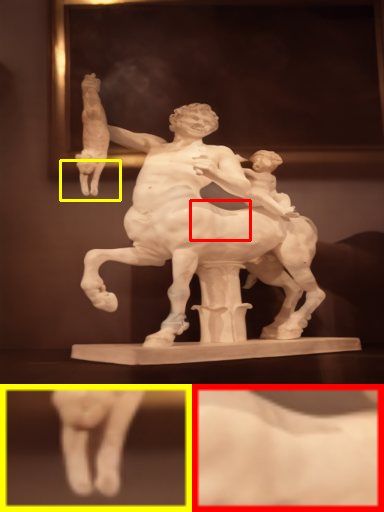}
		\vspace{-1.5em}
		\caption*{StableSR}
	\end{subfigure}
	\begin{subfigure}[c]{0.24\linewidth}
		\centering
		\includegraphics[width=0.88in,height=0.88in]{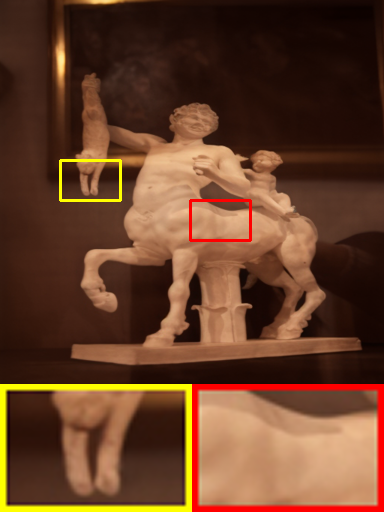}
		\vspace{-1.5em}
		\caption*{+Ours}
	\end{subfigure}
	\begin{subfigure}[c]{0.24\linewidth}
		\centering
		\includegraphics[width=0.88in,height=0.88in]{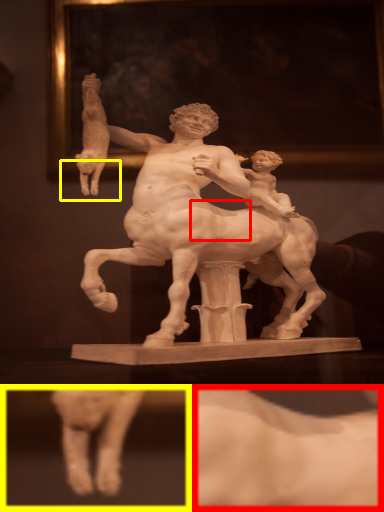}
		\vspace{-1.5em}
		\caption*{GT}
	\end{subfigure}

	\begin{subfigure}[c]{0.24\linewidth}
		\centering
		\includegraphics[width=0.88in,height=0.88in]{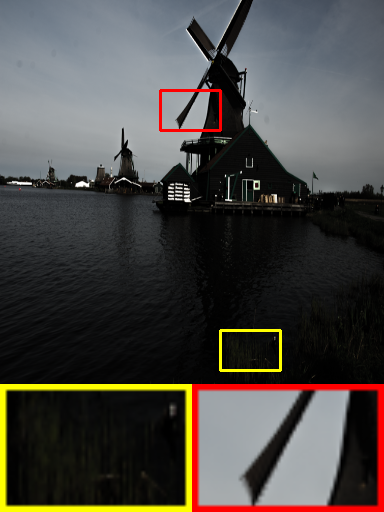}
		\vspace{-1.5em}
		\caption*{LOL-syn.}
	\end{subfigure}
	\begin{subfigure}[c]{0.24\linewidth}
		\centering
		\includegraphics[width=0.88in,height=0.88in]{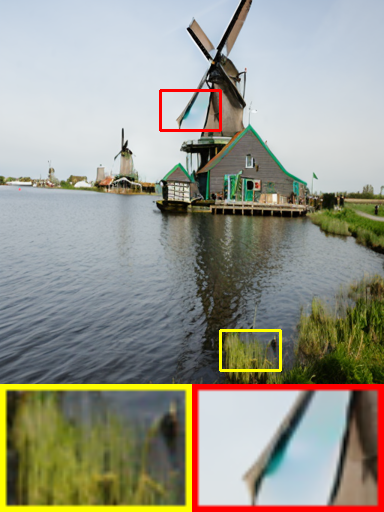}
		\vspace{-1.5em}
		\caption*{PASD}
	\end{subfigure}
	\begin{subfigure}[c]{0.24\linewidth}
		\centering
		\includegraphics[width=0.88in,height=0.88in]{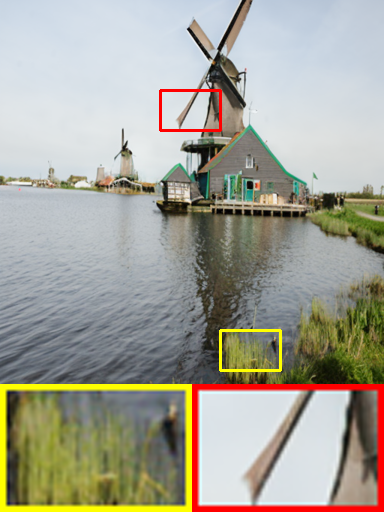}
		\vspace{-1.5em}
		\caption*{+Ours}
	\end{subfigure}
	\begin{subfigure}[c]{0.24\linewidth}
		\centering
		\includegraphics[width=0.88in,height=0.88in]{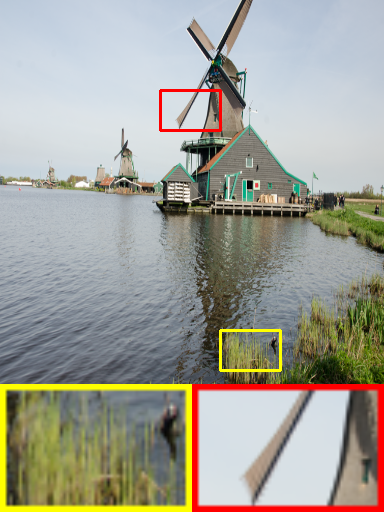}
		\vspace{-1.5em}
		\caption*{GT}
	\end{subfigure}

	\caption{Visual comparisons on different datasets with various network structures on LOL-synthetic. 
    ``LOL-syn.'' means LOL-synthetic.
    }
	\label{fig:cmp2}
\end{figure}

\begin{figure}[t]
\centering
	\begin{subfigure}[c]{0.24\linewidth}
		\centering
		\includegraphics[width=0.88in]{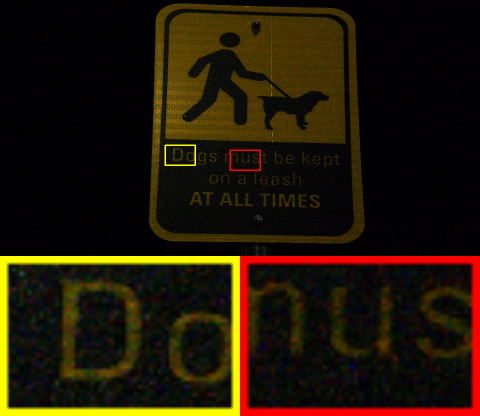}
		\vspace{-1.5em}
		\caption*{SID}
	\end{subfigure}
	\begin{subfigure}[c]{0.24\linewidth}
		\centering
		\includegraphics[width=0.88in]{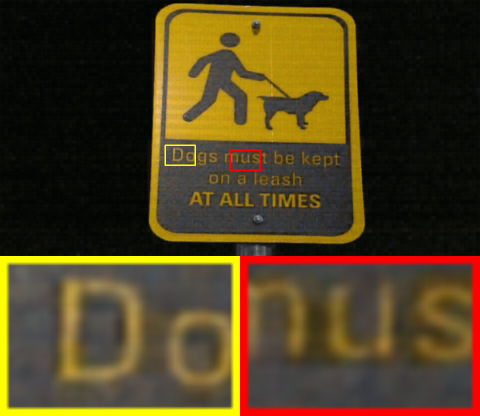}
		\vspace{-1.5em}
		\caption*{DiffBIR}
	\end{subfigure}
	\begin{subfigure}[c]{0.24\linewidth}
		\centering
		\includegraphics[width=0.88in]{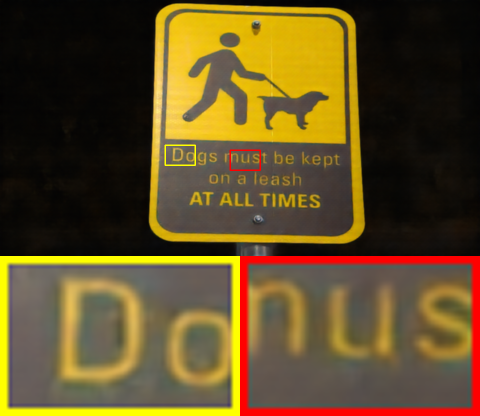}
		\vspace{-1.5em}
		\caption*{+Ours}
	\end{subfigure}
	\begin{subfigure}[c]{0.24\linewidth}
		\centering
		\includegraphics[width=0.88in]{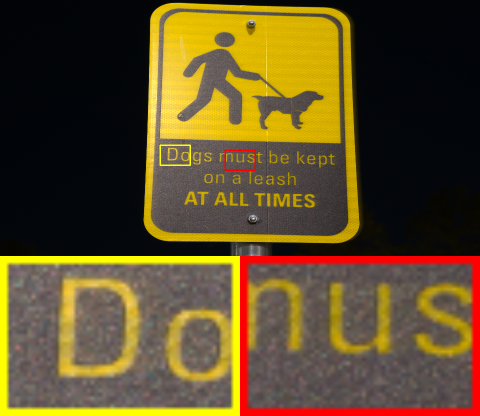}
		\vspace{-1.5em}
		\caption*{GT}
\end{subfigure}

	\begin{subfigure}[c]{0.24\linewidth}
		\centering
		\includegraphics[width=0.88in]{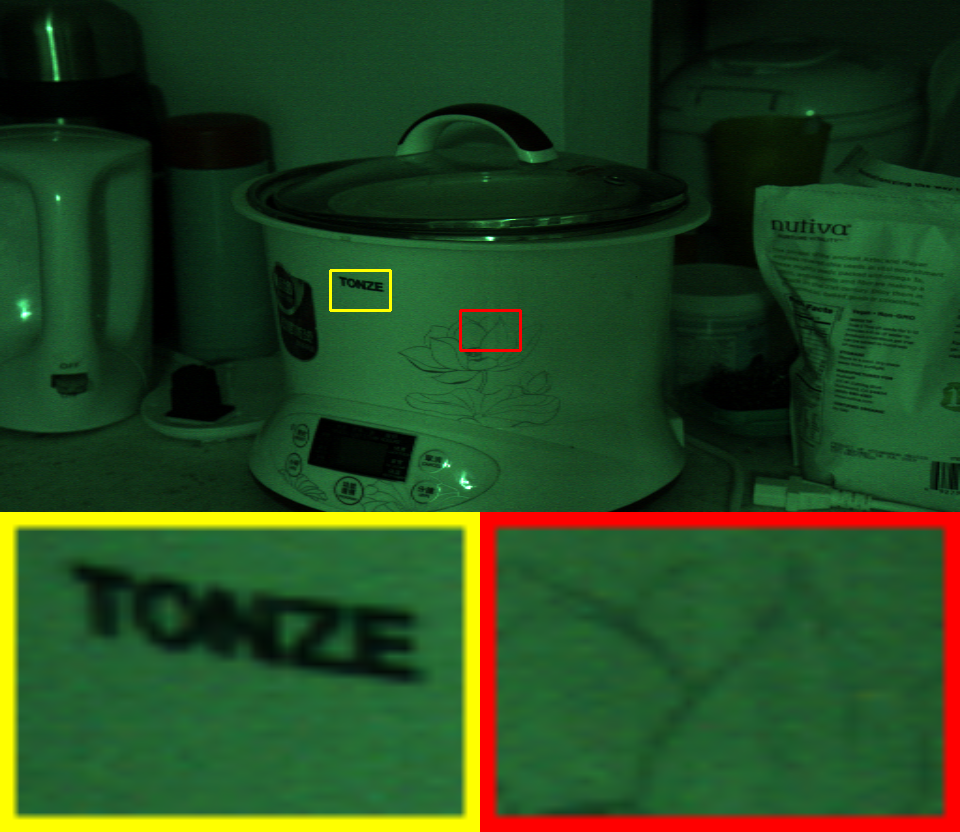}
		\vspace{-1.5em}
		\caption*{SID}
	\end{subfigure}
	\begin{subfigure}[c]{0.24\linewidth}
		\centering
		\includegraphics[width=0.88in]{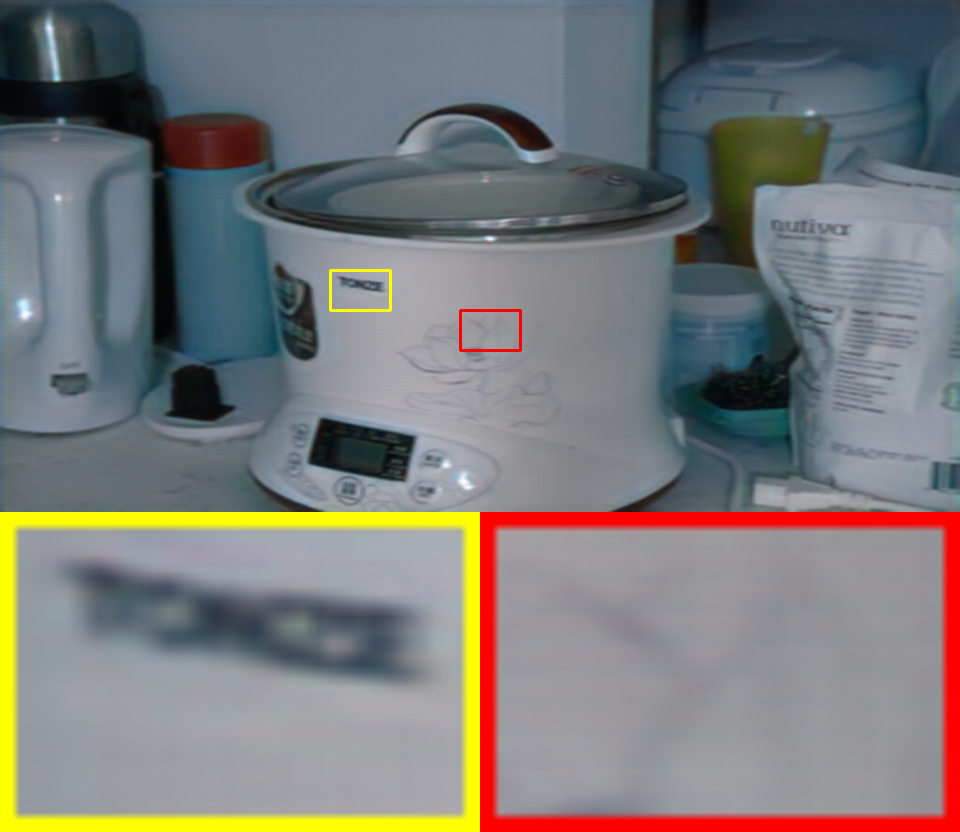}
		\vspace{-1.5em}
		\caption*{StableSR}
	\end{subfigure}
	\begin{subfigure}[c]{0.24\linewidth}
		\centering
		\includegraphics[width=0.88in]{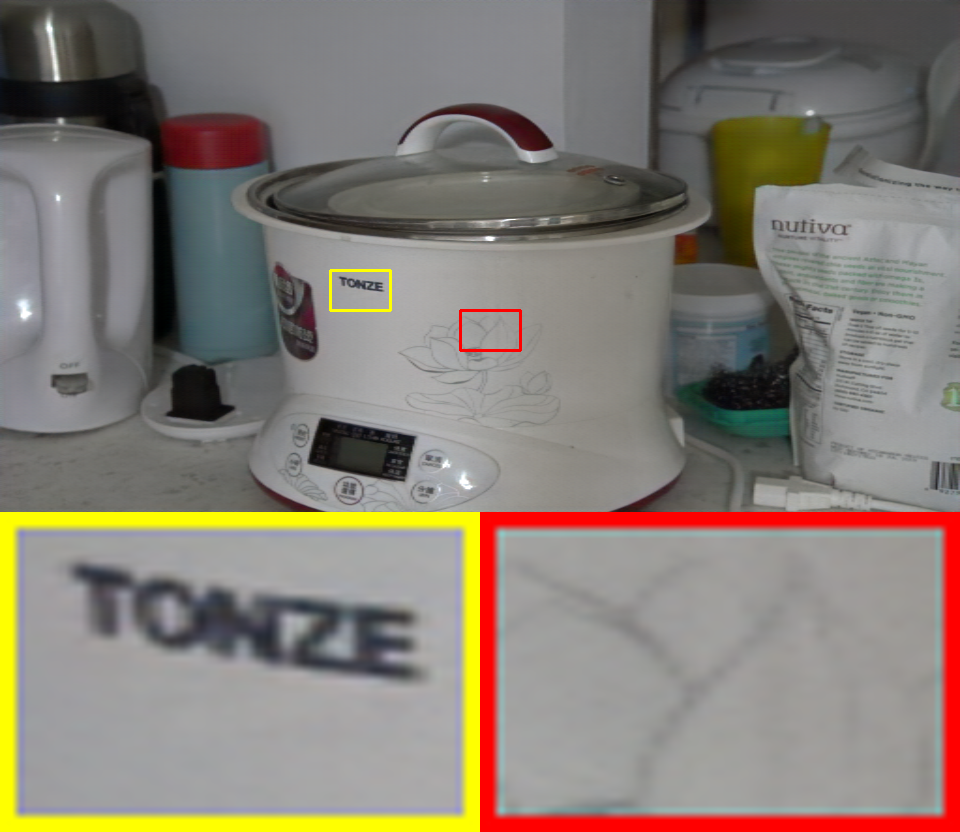}
		\vspace{-1.5em}
		\caption*{+Ours}
	\end{subfigure}
	\begin{subfigure}[c]{0.24\linewidth}
		\centering
		\includegraphics[width=0.88in]{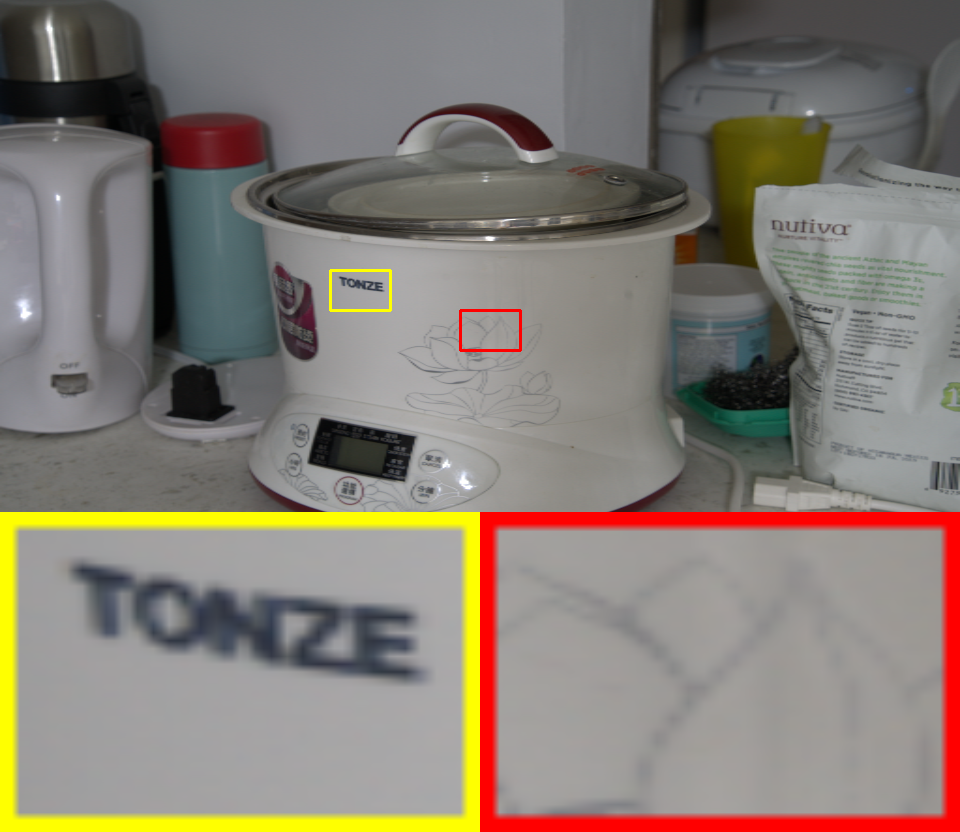}
		\vspace{-1.5em}
		\caption*{GT}
	\end{subfigure}

	\begin{subfigure}[c]{0.24\linewidth}
		\centering
		\includegraphics[width=0.88in]{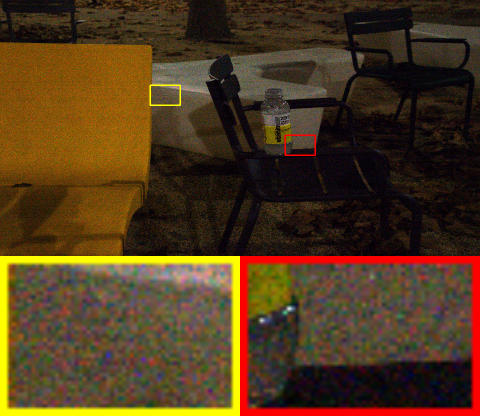}
		\vspace{-1.5em}
		\caption*{SID}
	\end{subfigure}
	\begin{subfigure}[c]{0.24\linewidth}
		\centering
		\includegraphics[width=0.88in]{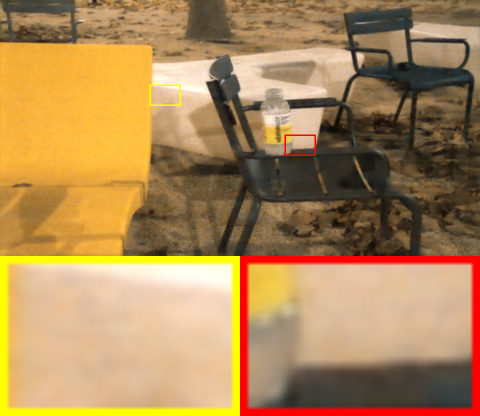}
		\vspace{-1.5em}
		\caption*{PASD}
	\end{subfigure}
	\begin{subfigure}[c]{0.24\linewidth}
		\centering
		\includegraphics[width=0.88in]{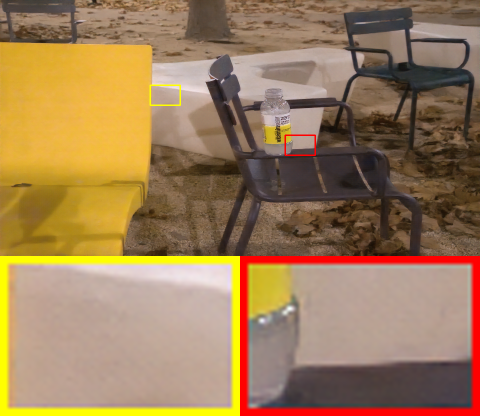}
		\vspace{-1.5em}
		\caption*{+Ours}
	\end{subfigure}
	\begin{subfigure}[c]{0.24\linewidth}
		\centering
		\includegraphics[width=0.88in]{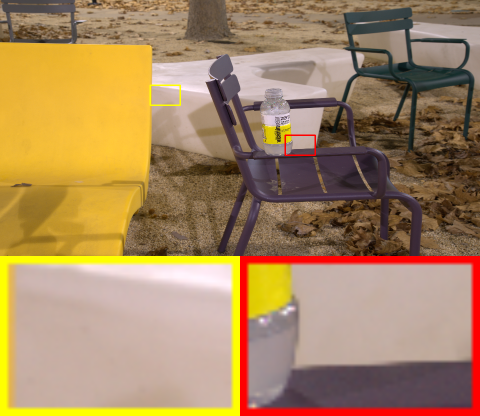}
		\vspace{-1.5em}
		\caption*{GT}
\end{subfigure}

\vspace{-0.1in}
	\caption{Visual comparisons on different datasets with various network structures on SID. 
    }
	\label{fig:cmp3}
\end{figure}

\begin{figure}[t]
\centering
\begin{subfigure}[c]{0.24\linewidth}
		\centering
		\includegraphics[width=0.88in]{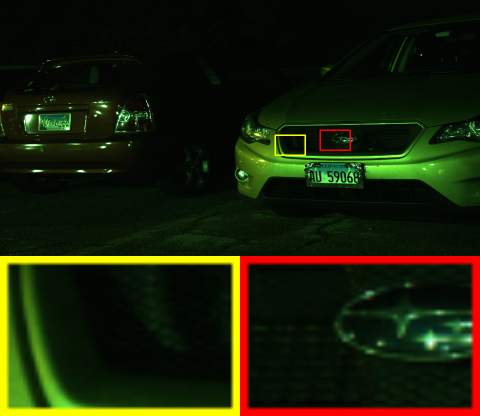}
		\vspace{-1.5em}
		\caption*{SMID}
	\end{subfigure}
	\begin{subfigure}[c]{0.24\linewidth}
		\centering
		\includegraphics[width=0.88in]{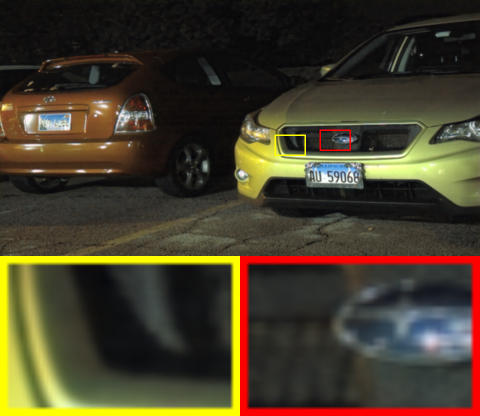}
		\vspace{-1.5em}
		\caption*{DiffBIR}
	\end{subfigure}
	\begin{subfigure}[c]{0.24\linewidth}
		\centering
		\includegraphics[width=0.88in]{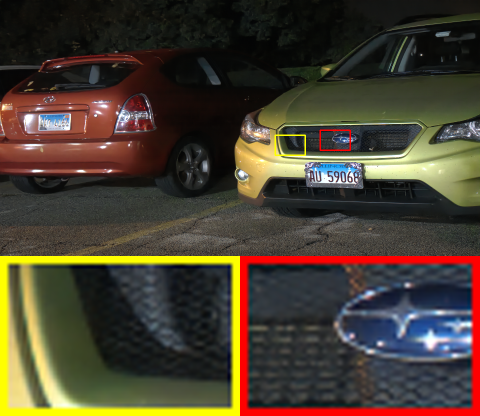}
		\vspace{-1.5em}
		\caption*{+Ours}
	\end{subfigure}
	\begin{subfigure}[c]{0.24\linewidth}
		\centering
		\includegraphics[width=0.88in]{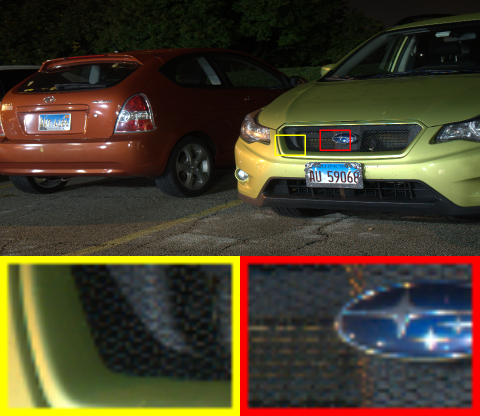}
		\vspace{-1.5em}
		\caption*{GT}
	\end{subfigure}

\begin{subfigure}[c]{0.24\linewidth}
		\centering
		\includegraphics[width=0.88in]{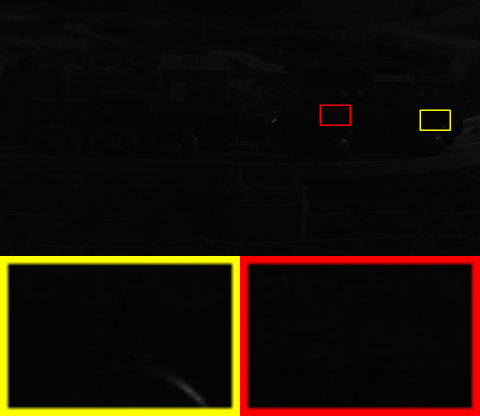}
		\vspace{-1.5em}
		\caption*{SMID}
	\end{subfigure}
	\begin{subfigure}[c]{0.24\linewidth}
		\centering
		\includegraphics[width=0.88in]{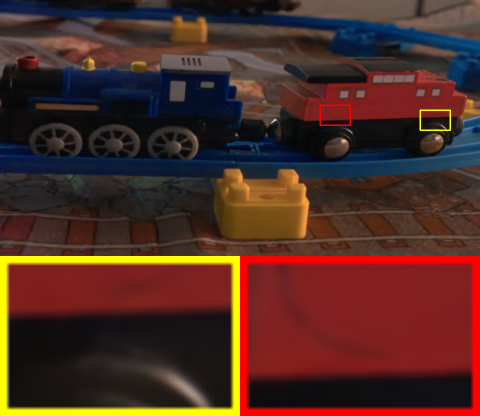}
		\vspace{-1.5em}
		\caption*{StableSR}
	\end{subfigure}
	\begin{subfigure}[c]{0.24\linewidth}
		\centering
		\includegraphics[width=0.88in]{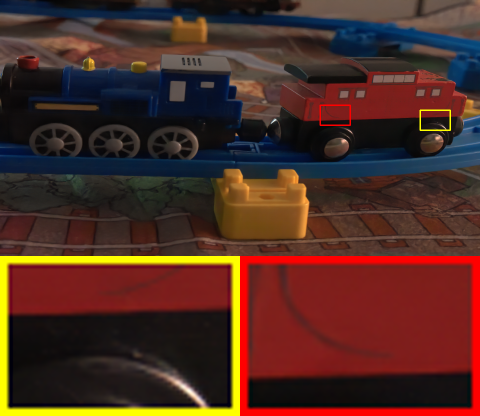}
		\vspace{-1.5em}
		\caption*{+Ours}
	\end{subfigure}
	\begin{subfigure}[c]{0.24\linewidth}
		\centering
		\includegraphics[width=0.88in]{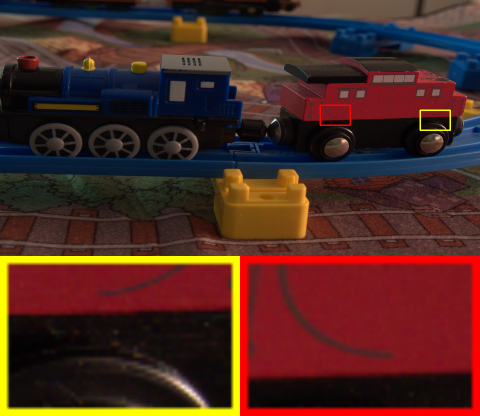}
		\vspace{-1.5em}
		\caption*{GT}
	\end{subfigure}

\begin{subfigure}[c]{0.24\linewidth}
		\centering
		\includegraphics[width=0.88in]{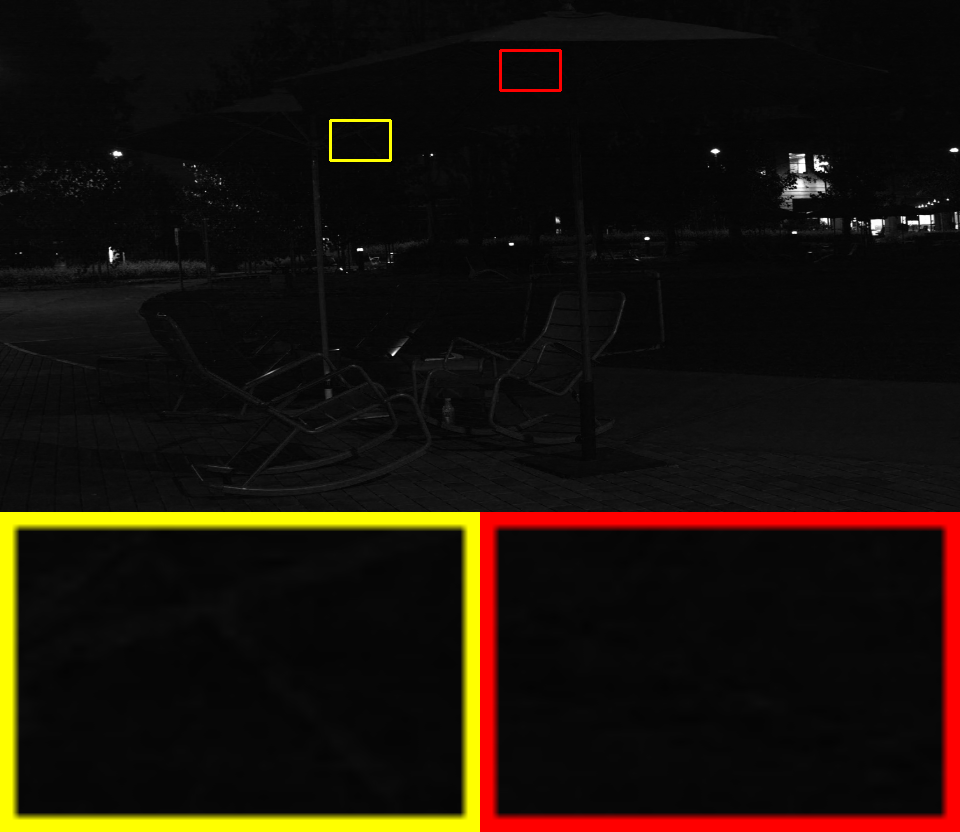}
		\vspace{-1.5em}
		\caption*{SMID}
	\end{subfigure}
	\begin{subfigure}[c]{0.24\linewidth}
		\centering
		\includegraphics[width=0.88in]{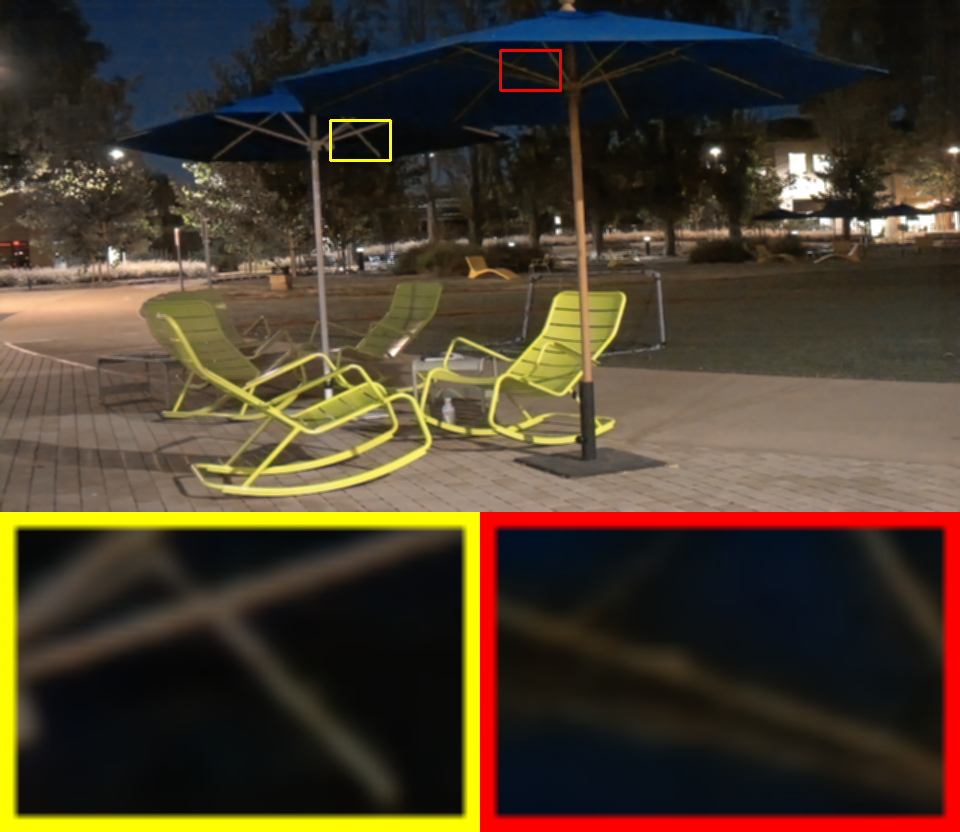}
		\vspace{-1.5em}
		\caption*{PASD}
	\end{subfigure}
	\begin{subfigure}[c]{0.24\linewidth}
		\centering
		\includegraphics[width=0.88in]{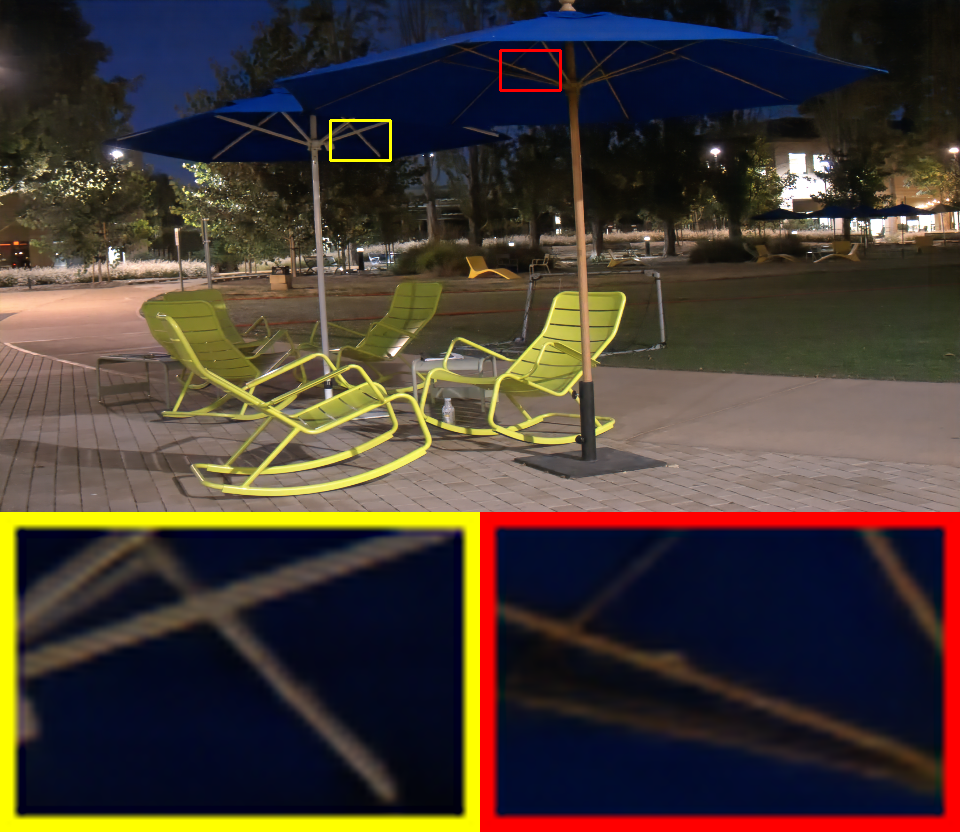}
		\vspace{-1.5em}
		\caption*{+Ours}
	\end{subfigure}
	\begin{subfigure}[c]{0.24\linewidth}
		\centering
		\includegraphics[width=0.88in]{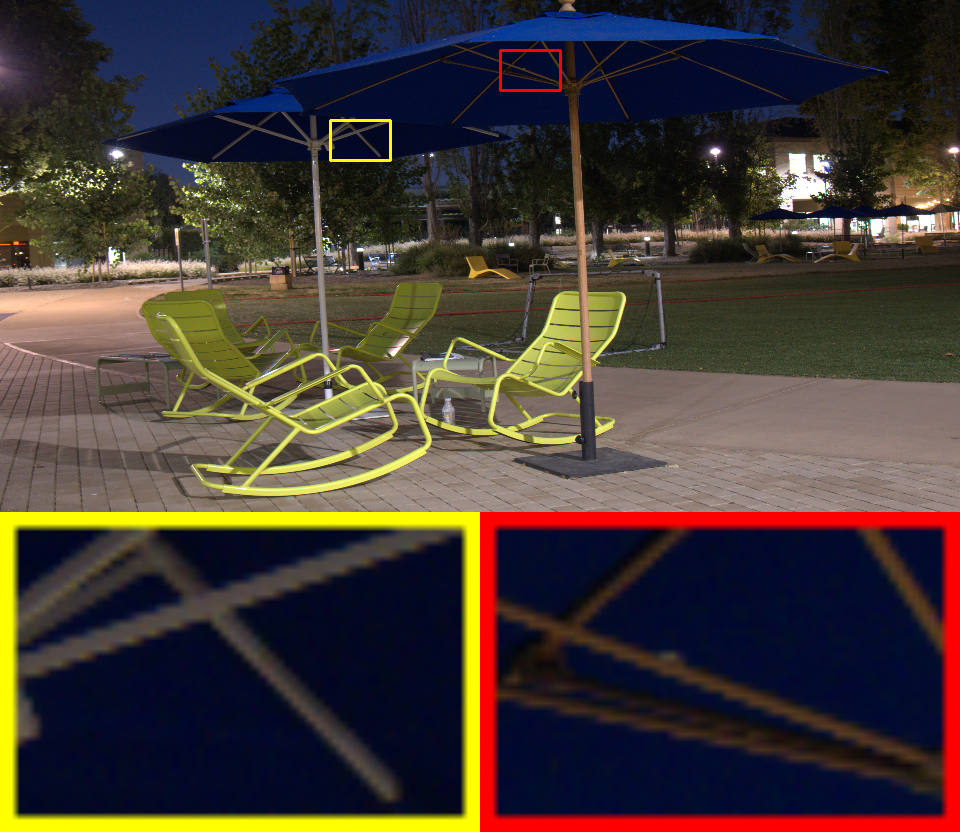}
		\vspace{-1.5em}
		\caption*{GT}
	\end{subfigure}
    \vspace{-0.1in}
	\caption{Visual comparisons on different datasets with various network structures on SMID. 
    }
	\label{fig:cmp4}
\end{figure}

\begin{table}[t]
	\centering
    \huge
	\resizebox{1.0\linewidth}{!}{
		\begin{tabular}{|l|p{2cm}<{\centering}p{2cm}<{\centering}p{2cm}<{\centering}p{2cm}<{\centering}|p{2cm}<{\centering}p{2cm}<{\centering}p{2cm}<{\centering}p{2cm}<{\centering}|}
            \hline
			& \multicolumn{4}{c|}{LOL-real} & \multicolumn{4}{c|}{LOL-synthetic}\\
			\hline
            Methods & PSNR$\uparrow$& SSIM$\uparrow$&LPIPS$\downarrow$&FID$\downarrow$& PSNR$\uparrow$& SSIM$\uparrow$&LPIPS$\downarrow$&FID$\downarrow$\\
			\hline \hline
			DiffBIR & 16.89 &0.717 &0.1139 &88.61 &20.25  &0.752 &0.1004 &40.17  \\
			DiffBIR +Ours & \textbf{20.28} &\textbf{0.746} &\textbf{0.0988} & \textbf{80.66}& \textbf{21.62} &\textbf{0.761} &\textbf{0.0716} & \textbf{34.99} \\ \hline
            StableSR & 20.39 &0.735 &0.1227 &76.71& 23.42 & 0.784& 0.1173&42.66   \\
            StableSR +Ours & \textbf{22.18} &\textbf{0.750} &\textbf{0.0964} &\textbf{73.15}&\textbf{24.50}  &\textbf{0.808} & \textbf{0.0941}& \textbf{40.65} \\ \hline
            PASD  & 20.58 & 0.729& 0.1095 &78.89 & 22.86 &0.780 &0.0935 &38.76 \\
            PASD +Ours  & \textbf{22.15} &\textbf{0.749} & \textbf{0.0953}&\textbf{75.64} & \textbf{24.27} &\textbf{0.803}& \textbf{0.0758}& \textbf{36.64} \\ 
            \hline

XPSR~\cite{qu2024xpsr} &21.15 &0.730 &0.1003 &75.47& 23.04 &0.786& 0.0918& 36.28  \\
XPSR+ours&\textbf{22.67} & \textbf{0.755}&  \textbf{0.0908}&  \textbf{72.74}&\textbf{24.03} &\textbf{0.791} &\textbf{0.0876} &\textbf{32.75}\\
\hline
TSD-SR~\cite{dong2024tsd}& 21.24& 0.737&0.1026 & 77.83& 23.15&0.769 &0.0954 &38.42\\
TSD-SR+ours&\textbf{22.72} &\textbf{0.756} &\textbf{0.0925} & \textbf{72.69}&\textbf{24.31} &\textbf{0.785} & \textbf{0.0903}&\textbf{35.07}\\
\hline
RAP~\cite{wang2024rap} &21.79 & 0.741&0.1042 &79.55 & 23.48&0.753 &0.0972 &39.50\\
RAP+ours&\textbf{22.81} &\textbf{0.763} &\textbf{0.0939} &\textbf{76.08} &\textbf{24.80} &\textbf{0.774} &\textbf{0.0867} &\textbf{36.49}\\
\hline
FaithDiff~\cite{chen2024faithdiff} &22.05 &0.749 &0.0934 &74.07 & 23.92&0.771 &0.0883 &35.61\\
FaithDiff+ours&\textbf{22.86} &\textbf{0.768} &\textbf{0.0890} &\textbf{70.21} &\textbf{24.67} & \textbf{0.782}& \textbf{0.0810}&\textbf{33.14}\\
\hline
Pixel~\cite{sun2024pixel} & 21.08& 0.724&0.0987 & 78.46&23.36 &0.750 &0.0975 &40.24\\
Pixel+ours&\textbf{22.50} &\textbf{0.743} & \textbf{0.0881}&\textbf{74.72} &\textbf{24.09} & \textbf{0.758}&\textbf{0.0901} &\textbf{37.96}\\
					\hline
	\end{tabular}}
    	\caption{Quantitative comparison between SOTA PTDB methods and their versions with our strategy on LOL-real and LOL-synthetic.
        }
	\label{comparison1}
\end{table}

\begin{table}[t]
	\centering
	\resizebox{\linewidth}{!}{
		\begin{tabular}{|l|p{0.7cm}<{\centering}p{0.7cm}<{\centering}p{0.7cm}<{\centering}p{0.7cm}<{\centering}|p{0.7cm}<{\centering}p{0.7cm}<{\centering}p{0.7cm}<{\centering}p{0.7cm}<{\centering}|}
            \hline
			& \multicolumn{4}{c|}{SID} & \multicolumn{4}{c|}{SMID}\\
			\hline
            Methods & PSNR$\uparrow$& SSIM$\uparrow$&LPIPS$\downarrow$&FID$\downarrow$& PSNR$\uparrow$& SSIM$\uparrow$&LPIPS$\downarrow$&FID$\downarrow$\\
			\hline \hline
			DiffBIR & 17.85 &0.604&0.2178 &90.62 & 22.47 &0.763 & 0.1836& 88.21\\
			DiffBIR +Ours  &\textbf{21.26}  & \textbf{0.622}& \textbf{0.1982}&\textbf{86.07} & \textbf{24.05} &\textbf{0.779} &\textbf{0.1631} &\textbf{85.18}  \\ \hline
            StableSR  & 20.27 &0.620 &0.2074 &87.39 & 24.08 &0.773 &0.1798 &85.75 \\
            StableSR +Ours & \textbf{21.48} &\textbf{0.651}& \textbf{0.1753}&\textbf{84.25} &\textbf{25.14} &\textbf{0.782} & \textbf{0.1537}& \textbf{82.76}\\ \hline
            PASD   & 20.62 &0.674& 0.1958 &81.83 &24.78  &0.780 &0.1856 &83.14 \\
            PASD +Ours  & \textbf{21.83} &\textbf{0.705} & \textbf{0.1770}&\textbf{78.14}&\textbf{25.42}  & \textbf{0.791} &\textbf{0.1704} &\textbf{80.37}\\ 
            \hline
	\end{tabular}}
    	\caption{The quantitative comparison between current SOTA PTDB methods and their versions with our strategy on SID and SMID.
        }
	\label{comparison1-1}
\end{table}

\begin{table*}[t]
    \footnotesize
	\centering
    {
        \begin{tabular}{|l|cccc|cccc|}
			\hline
			& \multicolumn{4}{c|}{LOL-real} & \multicolumn{4}{c|}{LOL-synthetic}\\
			\hline
			Methods & NIQE$\downarrow$& MUSIQ$\uparrow$&MANIQA$\uparrow$&CLIPIQA$\uparrow$& NIQE$\downarrow$& MUSIQ$\uparrow$&MANIQA$\uparrow$&CLIPIQA$\uparrow$\\
			\hline \hline
			DiffBIR & 6.7712&67.78&0.6034&0.6645&7.8546&69.11&0.6546&0.7187 \\
			DiffBIR +Ours &\textbf{6.5436}& \textbf{69.05}& \textbf{0.6392}& \textbf{0.6831}& \textbf{7.6838}& \textbf{70.63}& \textbf{0.6803}& \textbf{0.7309} \\ \hline
			StableSR&6.5214&65.27&0.5857&0.6428&7.6113&67.46&0.6372&0.6855\\
			StableSR +Ours &\textbf{6.2418}&\textbf{66.34}&\textbf{0.6139}&\textbf{0.6663}&\textbf{7.3531}&\textbf{68.48}&\textbf{0.6695}&\textbf{0.7030}  \\ \hline
			PASD  & 6.6705 & 63.09& 0.5778 &0.6284 &7.5597&66.47 &0.6521 &0.6683 \\
			PASD +Ours  & \textbf{6.4159} &\textbf{66.41} & \textbf{0.6076}&\textbf{0.6547} & \textbf{7.2540} &\textbf{67.75}& \textbf{0.6724}& \textbf{0.6922} \\ \hline
			\hline
			& \multicolumn{4}{c|}{SID} & \multicolumn{4}{c|}{SMID}\\
			\hline
			Methods & NIQE$\downarrow$& MUSIQ$\uparrow$&MANIQA$\uparrow$&CLIPIQA$\uparrow$&NIQE$\downarrow$& MUSIQ$\uparrow$&MANIQA$\uparrow$&CLIPIQA$\uparrow$\\
			\hline \hline
			DiffBIR  &4.5271&59.34&0.5527 &0.5443&5.7109&62.64&0.5838&0.6175\\
			DiffBIR +Ours&\textbf{4.2323} &\textbf{61.12}  &\textbf{0.5872}&\textbf{0.5724}&\textbf{5.4556}&\textbf{64.85}&\textbf{0.6061}&\textbf{0.6327}\\ \hline
			StableSR& 4.7031 &62.45& 0.5719&0.5577&5.8128&64.86&0.6180&0.6075\\
			StableSR +Ours& \textbf{4.5546} &\textbf{63.72}&\textbf{0.6008}&\textbf{0.5945}&\textbf{5.5867}&\textbf{66.34}&\textbf{0.6364}&\textbf{0.6280}\\ \hline
			PASD &4.5158 &60.81 &0.5633&0.5304&5.5701&61.49&0.6077&0.6212\\
			PASD +Ours &\textbf{4.2385} &\textbf{62.27} &\textbf{0.5884}&\textbf{0.5570}&\textbf{5.3422}&\textbf{63.12}&\textbf{0.6256}&\textbf{0.6503}\\ 
			\hline
	\end{tabular}}
    	\caption{The quantitative comparison between current SOTA methods and their versions with our strategy on different datasets, using no-reference image quality measures.}
	\label{comparison1-noref}
\end{table*}

\begin{table}[t]
	\centering
	\resizebox{1.0\linewidth}{!}{
		\begin{tabular}{|c|p{1cm}<{\centering}p{1.2cm}<{\centering}|p{1cm}<{\centering}p{1.4cm}<{\centering}|p{1cm}<{\centering}p{1.2cm}<{\centering}|}
			\hline
            Methods&Q.&Q.+Ours&L.D. &L.D.+Ours &Li.&Li.+Ours\\
            \hline
            PSNR &20.59 &\textbf{21.36} &19.95 & \textbf{20.84} &22.03 &\textbf{22.71} \\
            SSIM &0.811 &\textbf{0.820} & 0.781 & \textbf{0.792} &0.862 &\textbf{0.869} \\
            \hline
	\end{tabular}}
\caption{The quantitative comparison between current unsupervised SOTA PTDB methods (designed for low-light image enhancement) and their versions with our method on the LOL-real dataset. ``Q.", ``L.D.", and ``Li." denote QuadPrior, LLIEDiff, and LightenDiffusion, respectively.}
	\label{comparison1-unsupervised}
\end{table}

\begin{table}[t]
    \footnotesize
	\centering
	\resizebox{1.0\linewidth}{!}
    {
        \begin{tabular}{|c|cccc|}
			\hline
            Methods&Diff-L&Diff-Retinex&QuadPrior &GASD \\
            \hline
            PSNR$\uparrow$&21.45&21.81&20.56&20.28 \\
            SSIM$\uparrow$&0.571&0.695&0.629&0.653\\
            NIQE$\downarrow$ & 5.5466 & 5.2362&5.0159 & 6.3027\\
            MUSIQ$\uparrow$& 54.09 & 57.13 & 56.25 & 50.02\\
            MANIQA$\uparrow$& 0.5064 & 0.5290 & 0.5203 & 0.4809\\
            CLIPIQA$\uparrow$& 0.4808 & 0.5014 & 0.5090 & 0.4526\\
            \hline
            Methods &AnlightenDiff &LLFlow&PASD&PASD+Ours\\
            \hline
            PSNR$\uparrow$&21.07 &  21.72 & 20.62& \textbf{21.83}\\
            SSIM$\uparrow$& 0.680& 0.618 & 0.674& \textbf{0.705}\\
            NIEQ$\downarrow$ &5.8462&5.3083& 4.5158 & \textbf{4.2385}\\
            MUSIQ$\uparrow$&  52.14 & 56.36 &  60.81&\textbf{62.27}\\
            MANIQA$\uparrow$& 0.4921 & 0.5107 &  0.5633 & \textbf{0.5884}\\
            CLIPIQA$\uparrow$& 0.4725 &0.4912 &  0.5304 & \textbf{0.5570} \\
            \hline
	\end{tabular}}
    \caption{The comparison between our approach and current SOTA methods (mainly strategies besides PTDB methods) on the challenging SID dataset.}
	\label{comparison1-mul}
\end{table}

\subsection{Datasets}
We evaluate our framework on several datasets with noise in low-light image regions, including LOL-real, LOL-synthetic~\cite{yang2021sparse}, SID~\cite{chen2018learning}, and SMID~\cite{chen2019seeing}. LOL-real contains 689 low-/normal-light image pairs for training and 100 pairs for testing. LOL-synthetic was created by analyzing the illumination distribution in the RAW format. SID and SMID consist of short- and long-exposure image pairs. 
For SID, we use the subset captured with a Sony camera and follow the provided script to convert the low-light images from RAW to RGB using rawpy's default ISP. For SMID, we use the full images and also convert RAW data to RGB, as our work focuses on low-light image enhancement in the RGB domain. We split the training and testing data as in~\cite{chen2019seeing}. 

\textit{Note that in this paper, we primarily focus on the task of low-light enhancement within PTDB methods, as it poses significant challenges and demands in improving image fidelity. Moreover, low-light enhancement is a highly practical task, e.g., as discussed in Sec.~\ref{sec:downstream}. After evaluating our method on this task, we further observe that it can be effectively extended to other image restoration tasks. The corresponding dataset configurations are provided in Sec.~\ref{sec:sr-other}, highlighting the potential of our approach for broader applications and future research directions.}

\subsection{Implementation Details}

We apply our method to various approaches that utilize pre-trained diffusion models, refining their conditional latent and incorporating a bidirectional interaction mechanism. While these methods are primarily evaluated on SR, they can also be adapted for other restoration tasks, such as low-light image enhancement.
For the prediction of the final conditional latent variable $\hat{\boldsymbol{l}}_m$, we observe that each latent channel possesses distinct characteristics in addition to certain shared properties. To account for this, the network $\gamma$ comprises a shared component for channel-wise prediction, followed by four specialized sub-networks, each dedicated to predicting the corresponding channel's output.  
Moreover, in our framework, there are four learnable modules besides the control module (e.g., ControlNet): the visual encoder $E_v$ for extracting $\boldsymbol{t}_l$, the lightweight latent diffusion model $\epsilon_\theta$, the final latent predictor $\gamma$, and the latent interaction module $\beta$. These modules integrate a combination of CNN and transformer architectures (i.e., SCUNet~\cite{zhang2023practical}, with a head dimension of 32, a window size of 8, and two blocks per layer), except in the ablation study setting of ``with Simple $\epsilon_\theta$/$\gamma$''.

Using DiffBIR as an example, the architecture details of each module are summarized as follows: the visual encoder $E_v$ in Table~\ref{tab:visual-encoder}, the lightweight latent diffusion $\epsilon_\theta$  in Tables~\ref{tab:denoise-diffusion} and \ref{tab:condition-diffusion}, the final latent predictor $\gamma$ in Tables~\ref{tab:prediction} and \ref{tab:prediction2}, and the latent interaction module $\beta$ in Tables~\ref{tab:trans} and \ref{tab:trans2}. In the `with Simple $\epsilon_\theta$/$\gamma$'' setting, all SCUNet blocks are removed.

The experiments are conducted using the officially released code, with identical hyperparameters (we adopt the original settings of hyperparameters, including the setting of randomness) and training epochs for both the baseline and our method (for fair and accurate comparisons). 
All experiments run on an A100 GPU with 80G memory under Ubuntu system.
Moreover, the experimental score for each method is typically reported as the average of three runs to ensure the statistical significance of the improvements.

The pre-trained restoration model, specifically the SNR-aware network~\cite{xu2022snr}, is utilized as a representative low-light image enhancement network. Similarly, the pre-trained diffusion model is Stable Diffusion, except in the ablation study, where we evaluate effects of different pre-trained restoration and diffusion models.
\textit{Note that our code will be released upon the publication}.

\subsection{Comparisons}

\noindent\textbf{Performance Comparison.}
Our baselines include representative diffusion-based methods.
We select publicly available methods, including StableSR~\cite{wang2024exploiting}, PASD~\cite{yang2024pixel}, and DiffBIR~\cite{lin2024diffbir}.
The comparison results are shown in Tables~\ref{comparison1} and \ref{comparison1-1}. As indicated, our strategy enhances the performance of various PTDB methods, improving both PSNR (fidelity) and SSIM (more related with image details). Notably, PTDB methods combined with our approach demonstrate significant gains, such as a 3.4 dB improvement in PSNR and a 0.3 increase in SSIM for DiffBIR on LOL-real. 

We also extend the evaluation to no-reference image quality measures, including NIQE~\cite{zhang2015feature}, MANIQA~\cite{yang2022maniqa}, MUSIQ~\cite{ke2021musiq}, and CLIPIQA~\cite{wang2023exploring}. The results, presented in Table~\ref{comparison1-noref}, demonstrate that our method continues to achieve the best performance.

Additionally, we provide visual examples to highlight the perceptual improvements. As shown in Figs.~\ref{fig:cmp}, \ref{fig:cmp2}, \ref{fig:cmp3}, and \ref{fig:cmp4}, the results enhanced by our method are closer to the ground truth.
Notably, our method produces results with higher fidelity. For example, in the visual comparison on LOL-real (Fig.~\ref{fig:cmp}), the baseline fails to accurately synthesize the colors and shapes of the stairs, handrail, and door. On LOL-synthetic, the baseline introduces artificial cloud formations, whereas our method preserves the original shape, better aligning with user expectations. Similarly, results on SID and SMID further support this conclusion: while the baseline fails to retain textual and other fine textures on SID, ours preserves and enhances them.
\textbf{More results can be found in the supplementary file}.

In addition, we apply our method to existing pre-trained diffusion-based low-light enhancement models, although they are primarily designed for unsupervised settings. Specifically, we refine the latent conditions using our strategy for QuadPrior~\cite{wang2024zero}, LLIEDiff~\cite{huang2025zero}, and LightenDiffusion~\cite{jiang2024lightendiffusion}. As shown in Table~\ref{comparison1-unsupervised}, performance is still improved, as richer details are introduced in the latent space.

\vspace{2mm}
\noindent\textbf{Comparison besides PTDB Methods.} Moreover, as shown in Table~\ref{comparison1-mul}, the baseline model PASD when integrated with our strategy can reach state-of-the-art performance on the dark, noisy, and challenging SID dataset, particularly in terms of no-reference image quality metrics.
The baselines include both diffusion-based approaches (without generative priors from the large pre-trained models, including Diff-L~\cite{jiang2023low}, Diff-Retinex~\cite{yi2023diff}, GASD~\cite{hou2024global}, and AnlightenDiff~\cite{chan2024anlightendiff}) and other restoration network architectures (e.g., LLFlow~\cite{wang2022low}). 
These baselines are both representative and recent, showing the performance potential of current restoration-only methods on SID.
These results in Table~\ref{comparison1-mul} further highlight the fidelity improvement brought by our proposed strategy.

\vspace{2mm}
\noindent\textbf{Efficiency.}
Furthermore, we analyze the efficiency of our method. Compared to the baseline, it requires only a small number of additional parameters. For example, when applied to the DiffBIR model, our method introduces just 0.05B additional parameters (significantly less than the original 1.46B parameters) resulting in a relative increase of only 3.4\%.
In addition, the runtime of our method is slightly higher than that of the baseline. Further efficiency optimization will be explored in future work.

\begin{table}[t]
	\centering
	\huge
    \resizebox{1.0\linewidth}{!}{
		\begin{tabular}{|l|p{1.8cm}<{\centering}p{1.8cm}<{\centering}p{1.8cm}<{\centering}p{1.8cm}<{\centering}|p{1.8cm}<{\centering}p{1.8cm}<{\centering}p{1.8cm}<{\centering}p{1.8cm}<{\centering}|}
            \hline
			& \multicolumn{4}{c|}{LOL-real} & \multicolumn{4}{c|}{LOL-synthetic}\\
			\hline
            Methods & PSNR$\uparrow$& SSIM$\uparrow$&LPIPS$\downarrow$&FID$\downarrow$& PSNR$\uparrow$& SSIM$\uparrow$&LPIPS$\downarrow$&FID$\downarrow$\\
			\hline \hline
			w/o $\boldsymbol{t}_i$ & 18.41 & 0.729& 0.1112&84.20 & 20.37 &0.740&0.0863 &38.35  \\
            w/o $\boldsymbol{t}_l$ & 16.87 &0.705 &0.1176 & 82.81&20.33  &0.734&0.0828 & 37.94 \\
            w/o prior & 19.43 &0.726 &0.1124 &82.48 & 20.70 &0.735& 0.0864& 38.53 \\
            w/o gen. & 17.17 &0.711 &0.1073 & 82.09& 20.61 &0.736&0.0827 &37.16  \\
            w/o pyramid  & 20.15 & 0.725 &0.1081 &87.83 & 20.42 &0.740&0.0755 &36.82  \\
            w/o interact &18.16  &0.718&0.1028 &86.97 & 19.76 & 0.709&0.0767 &37.41 \\
            w/o PL &17.75  &0.712 &0.1090 &85.84 & 20.98 &0.747&0.0782 & 38.59 \\
            w/o att. & 19.37 &0.739 &0.1039 &86.51 & 20.28 & 0.723& 0.0804&35.81 \\
            \hline
            with Restormer & 20.12 &0.735 &0.1067 & 82.95&20.84  &0.752&0.0841 & 37.76 \\
            with Simple $\epsilon_\theta$/$\gamma$ & 19.73& 0.729& 0.1104&83.72 & 21.10&0.758 &0.0795 &36.77\\
			Full & \textbf{20.28} &\textbf{0.746} &\textbf{0.0988} & \textbf{80.66}&\textbf{21.62}  &\textbf{0.761} &\textbf{0.0716} &\textbf{34.99} \\
            \hline
	\end{tabular}}

        	\caption{The ablation study results. The experiments are conducted with DiffBIR and SNR-aware network as the pretrained diffusion and restoration models, except ``with Restormer''.}
	\label{comparison1-abla}
\end{table}

\begin{table}[t]
	\centering
	\huge
    \resizebox{1.0\linewidth}{!}{
		\begin{tabular}{|l|p{1.8cm}<{\centering}p{1.8cm}<{\centering}p{1.8cm}<{\centering}p{1.8cm}<{\centering}|p{1.8cm}<{\centering}p{1.8cm}<{\centering}p{1.8cm}<{\centering}p{1.8cm}<{\centering}|}
            \hline
			& \multicolumn{4}{c|}{LOL-real} & \multicolumn{4}{c|}{LOL-synthetic}\\
			\hline
            Methods & PSNR$\uparrow$& SSIM$\uparrow$&LPIPS$\downarrow$&FID$\downarrow$& PSNR$\uparrow$& SSIM$\uparrow$&LPIPS$\downarrow$&FID$\downarrow$\\
			\hline \hline
            with Flux$_b$ & 19.04& 0.730 &0.1082 &84.61 & 20.82 & 0.759&0.0914 & 37.48 \\
            with Flux &\textbf{23.17}  & \textbf{0.783}&\textbf{0.0951} &\textbf{76.12} & \textbf{25.01} & \textbf{0.790}&\textbf{0.0657} & \textbf{32.93} \\
			with SD & \underline{20.28} &\underline{0.746} &\underline{0.0988} & \underline{80.66}&\underline{21.62}  &\underline{0.761} &\underline{0.0716} &\underline{34.99}  \\
            \hline
	\end{tabular}}

	\caption{The ablation study results with different pre-trained diffusion models. ``with Flux$_b$'' refers to the baseline (DiffBIR) using Flux \textit{without} our approach; ``with Flux''/``with SD" denote the baseline using Flux/Stable Diffusion \textit{with} our strategy.
    }
	\label{comparison1-abla2}
\end{table}

\subsection{Ablation Study}
Here, we present several ablation studies to analyze the impact of our proposed different strategies. 

\vspace{2mm}
\noindent\textbf{The effects of $\boldsymbol{t}_i$ and $\boldsymbol{t}_l$.}
In our conditional latent refinement strategy, we emphasize the importance of leveraging the original image's information ($\boldsymbol{t}_i$ and $\boldsymbol{t}_l$) which retain spatial details uncompressed by the VAE encoder. 
To validate this, we conduct ablation experiments by removing $\boldsymbol{t}_i$ and $\boldsymbol{t}_l$ from the diffusion process (Eq.~\eqref{eq:eps}) and the final prediction procedure (Eq.~\eqref{eq:lm}), respectively. These settings are denoted as ``w/o $\boldsymbol{t}_i$'' and ``w/o $\boldsymbol{t}_l$''.
As shown in Table~\ref{comparison1-abla}, performance decreases when $\boldsymbol{t}_i$ and $\boldsymbol{t}_l$ are removed, highlighting the importance of these lossless inputs.

\vspace{2mm}
\noindent\textbf{The results without diffusion-based priors.}
In this paper, we highlight the impact of using a generative approach to obtain a suitable prior ($\hat{\boldsymbol{h}}_{m,0}$ in Eq.~\ref{eq:refine1}), which can guide the refined condition toward its ground truth. This approach helps mitigate the ill-posed nature of the prediction. To validate its effectiveness, we design two ablation settings: (1) removing the diffusion component entirely, i.e., eliminating the prior (``w/o prior''), and (2) replacing the generative diffusion process with a regression network with the same capacity (``w/o gen.'').
The results in Table~\ref{comparison1-abla} confirm our assumptions, demonstrating that models with a generative component are more effective for conditional latent refinement.

\vspace{2mm}
\noindent\textbf{The removal of pyramidal prior shape.}
Constructing an effective prior is challenging. To address this, we design the prior $\hat{\boldsymbol{h}}_{m,0}$ in a pyramidal shape (Eq.~\ref{eq:pyramid}), incorporating multi-scale information of the target. To assess its impact, we conduct an ablation study by replacing the pyramidal prior with a single-scale target, i.e., setting the prior target directly to $l_m$. This ablation setting is referred to as ``w/o pyramid''. The comparison between ``w/o pyramid'' and ``Full'' in Table~\ref{comparison1-abla} illustrates the impact of this design.

\vspace{2mm}
\noindent\textbf{The effects of removing interaction.}
We highlight the importance of the bidirectional interaction mechanism (Eqs.~\ref{eq:inter1} and \ref{eq:inter2} in Sec.~\ref{sec:inter}), as the noisy latent and conditional latent provide complementary information, enhancing control capability. To validate its effect, we conduct an ablation study by removing the bidirectional interaction, denoted as ``w/o interact''.
The importance of the interaction is verified by the comparison between ``w/o interact'' and ``Full'' in Table~\ref{comparison1-abla}.

\vspace{2mm}
\noindent\textbf{The effects of removing pixel-level loss.}
In this ablation study, we evaluate the effect of the loss function in Eq.~\eqref{eq:refine2}, which penalizes pixel-level inconsistencies between the refined conditional latent and the ground truth. To assess its impact, we remove Eq.~\eqref{eq:refine2} from the training process, referring to this setting as ``w/o PL''. The results in Table~\ref{comparison1-abla} support the role of pixel-level supervision.

\vspace{2mm}
\noindent\textbf{The effects of attention-aware prediction.}
In Eq.~\eqref{eq:lm}, we adopt an attention-aware prediction manner, focusing on the refinement of unsatisfied areas in $\boldsymbol{l}_c$.
In this setting, we remove the output of $\hat{\boldsymbol{l}}_{w,m}$.
This setting is called ``w/o attention'', and its effect is demonstrated by the results in Table~\ref{comparison1-abla}.

\vspace{2mm}
\noindent\textbf{The effects for different pre-trained diffusion models.}
With the advancement of diffusion models, various pre-trained backbones have emerged. To evaluate their impact, we conduct experiments using different pre-trained diffusion backbones, including the advanced Flux~\cite{flux2024}, while keeping other components unchanged. Results in Table~\ref{comparison1-abla2} show that our strategy (``with Flux'') performs better when combined with an advanced diffusion backbone.

\begin{table*}[t]
    \footnotesize
	\centering
    \begin{tabular}{|l|cc|cc|cc|}
            \hline
			Methods & DiffBIR&+Ours & StableSR &+Ours& PASD &+Ours \\
			\hline
			Top-1 (\%) on CODaN~$\textcolor{black}{\uparrow}$  &53.17 &\textbf{55.36} & 55.19& \textbf{57.80}& 56.82&\textbf{59.01} \\
			mIoU on Nighttime Driving~$\textcolor{black}{\uparrow}$ &23.8&\textbf{26.0} &26.7 & \textbf{29.1}&27.3& \textbf{29.5} \\
			mIoU on Dark-Zurich~$\textcolor{black}{\uparrow}$&22.3& \textbf{25.7}&25.6&\textbf{27.2}&25.9&\textbf{28.4}  \\
            \hline
            ACC on DARK FACE~$\textcolor{black}{\uparrow}$ &0.635 &\textbf{0.674} &0.646 &\textbf{0.682} &0.641 &\textbf{0.687}\\
            ACC on RealCE~$\textcolor{black}{\uparrow}$ &0.874 &\textbf{0.901} & 0.890&\textbf{0.908} &0.865 &\textbf{0.892}\\
            NED on RealCE~$\textcolor{black}{\uparrow}$ &0.882 &\textbf{0.897} &0.874 &\textbf{0.886} & 0.863&\textbf{0.879}\\
            \hline
	\end{tabular}
     	\caption{Quantitative comparisons on different downstream tasks.}
	\label{comparison4}
\end{table*}

\begin{table}[t]
	\centering
	\resizebox{1.0\linewidth}{!}{
		\begin{tabular}{|l|p{1cm}<{\centering}p{1cm}<{\centering}p{1cm}<{\centering}p{1.2cm}<{\centering}p{1.2cm}<{\centering}p{1.4cm}<{\centering}|}
			\hline
			Methods& PSNR$\uparrow$ & SSIM$\uparrow$  & LPIPS$\downarrow$  &MANIQA$\uparrow$  &MUSIQ$\uparrow$  &CLIP-IQA$\uparrow$  \\
			\hline \hline
			DiffBIR &21.9154&0.4986& 0.4263&61.1476&0.2466&0.6347\\
			+Ours &\textbf{22.5510}&\textbf{0.5094}& \textbf{0.4015}&\textbf{64.4539}&\textbf{0.2928}&\textbf{0.6710}\\
			\hline
			StableSR&21.2392&0.4790&0.3993&57.8069&0.1648&0.5541\\
			+Ours &\textbf{22.1412}&\textbf{0.4873}&\textbf{0.3754}&\textbf{61.1538}&\textbf{0.1905}&\textbf{0.5948}\\
			\hline
			PASD &20.7838&0.4727&0.4353& 63.8094&0.2354&0.6125\\
			+Ours &\textbf{21.6201}&\textbf{0.4962}&\textbf{0.4007}& \textbf{66.9502}&\textbf{0.2703}&\textbf{0.6618}\\
			\hline
	\end{tabular}}
        	\caption{The comparisons on the DIV2K-Val dataset.}
	\label{comparison1-sr}
\end{table}

\begin{table}[t]
	\centering
	\resizebox{1.0\linewidth}{!}{
		\begin{tabular}{|l|p{2.5cm}<{\centering}|p{2.5cm}<{\centering}|p{2.5cm}<{\centering}|}
			\hline
			Methods&MANIQA$\uparrow$  &MUSIQ$\uparrow$  &CLIP-IQA$\uparrow$  \\
			\hline \hline
			DiffBIR &69.4208&0.3211&0.7637\\
			+Ours &\textbf{70.3687}&\textbf{0.3482}&\textbf{0.7854}\\
			\hline
			StableSR&64.8372&0.2083&0.6418\\
			+Ours & \textbf{66.2631} & \textbf{0.2417} & \textbf{0.6705}\\
			\hline
			PASD & 67.4052&0.2370&0.6761\\
			+Ours & \textbf{68.6015} & \textbf{0.2593} & \textbf{0.6921}\\
			\hline
	\end{tabular}}
        	\caption{The comparisons on the RealSRSet~\cite{zhang2021designing}.}
	\label{comparison2-sr}
    
\end{table}

\vspace{2mm}
\noindent\textbf{The effects for different pre-trained restoration models.}
We also assess the impact of incorporating different pre-trained restoration models within our pipeline to obtain $\boldsymbol{I}_{d_i}$ from $\boldsymbol{I}_{d_l}$. Specifically, we replace the SNR-aware network with Restormer~\cite{zamir2022restormer} to evaluate the effect of this change. Generally, the SNR-aware network outperforms Restormer in terms of restoration quality.  
The results (``with Restormer''), presented in Table~\ref{comparison1-abla}, indicate that our approach remains competitive even when using Restormer instead of the SNR-aware network. This shows the effectiveness of our latent refinement strategy, ensuring that the final conditional latent \(\hat{\boldsymbol{l}}_m\) mitigates the dependency on the pre-trained restoration model.

\vspace{2mm}
\noindent\textbf{Influence of network capacity of $\epsilon_\theta$ and $\gamma$.}
$\epsilon_\theta$ and $\gamma$ are key components for latent refinement (Eqs.~\eqref{eq:eps} and \eqref{eq:lm}).
We investigate whether the effectiveness of latent refinement stems from the large network capacity of these components, which combine CNN and transformer architectures in this work. To explore this, we conduct experiments by setting $\epsilon_\theta$ and $\gamma$ as small CNN networks. The results, presented in Table~\ref{comparison1-abla} under ``with Simple $\epsilon_\theta$/$\gamma$'', show that, although performance decreases compared to ``Full'', it still outperforms the SOTA baseline in Table~\ref{comparison1}. This indicates that our latent refinement's impact is not solely due to additional learnable parameters (as also supported by comparisons among ``w/o prior'', ``w/o gen.'', and ``Full''), but rather the effective modeling strategy.

\subsection{User Study}
To demonstrate the visual improvements brought by our method, we conducted a user study with 20 participants, focusing on the low-light image enhancement task. We randomly selected 30 images from the test sets of LOL, SID, and SMID for evaluation. Following common practice in low-light enhancement studies, we adopted an AB-test protocol.
In each comparison, the result produced by our method is labeled as ``Image A'', while the baseline result is labeled as ``Image B''. During the evaluation, participants were shown both images simultaneously in a randomized left-right order to avoid positional bias. Each participant compared the outputs of our method and the baselines in a random order across 30 tasks.
Participants were asked to select one of three options: ``Image A is better'', ``Image B is better'', or ``I think they are of the same quality''. Their judgments were based on criteria such as natural brightness, contrast, color fidelity, detail richness, and artifact reduction.

\begin{figure}[t]
	\centering
	\includegraphics[width=.88\linewidth]{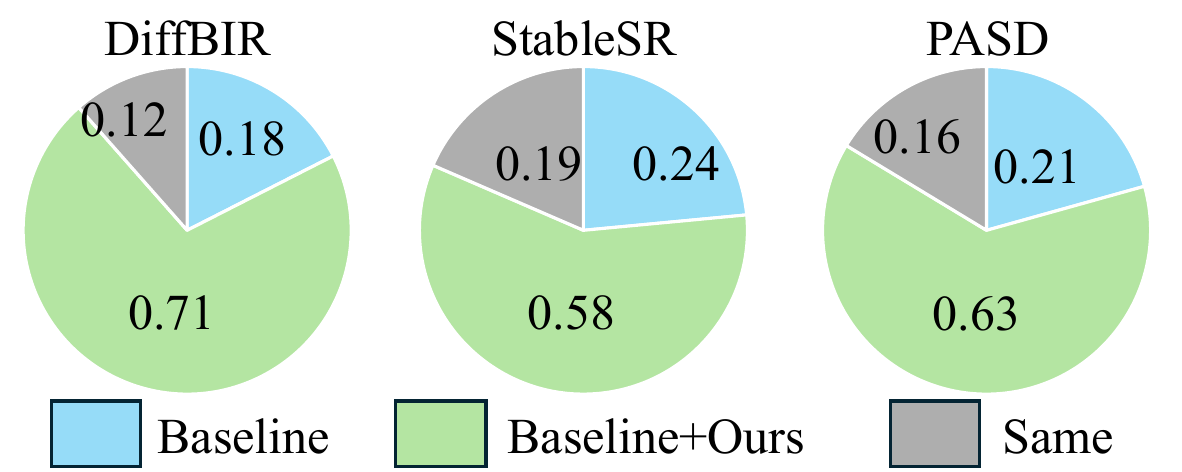}
	\caption{
		The above pie charts summarize the results of our user study. It is evident that the results enhanced with our strategy are preferred by the participants.
	}
	\label{fig:user}
\end{figure}

Fig.~\ref{fig:user} summarizes the user study's results, and we can see that ours gets more selections from participants over the baselines. This demonstrates that our method's results are more preferred by the human subjective perception.

\subsection{The Evaluation with Downstream Tasks}
\label{sec:downstream}
Low-light image enhancement can improve the accuracy of downstream applications, e.g., help autonomous vehicles in nighttime driving.
We first evaluate two downstream tasks: the image classification and semantic segmentation. 
For image classification, we use CODaN~\cite{lengyel2021zero}, a 10-class dataset with daytime training images and test images that include both daytime and nighttime scenes (the backbone is
ResNet-18~\cite{he2016deep}). For semantic segmentation, we use two datasets: Nighttime Driving~\cite{dai2018dark} and Dark-Zurich~\cite{sakaridis2019guided}, which contain 50 coarsely annotated and 151 densely annotated nighttime street view images. The segmentation network is RefineNet~\cite{lin2017refinenet} with ResNet-101 backbone.
Table~\ref{comparison4} presents the evaluation results. The improvements achieved by our method across these tasks demonstrate its effectiveness in enhancing downstream applications.

Moreover, we focus on downstream tasks that demand high fidelity for human perception, such as face and text recognition.
For face recognition, we use S3FD~\cite{zhang2017s3fd}, a well-known face detection algorithm, to evaluate face detection performance on the DARK FACE dataset~\cite{yang2020advancing}. To assess text fidelity in restored text images, we employ word accuracy (ACC) and normalized edit distance (NED)~\cite{ma2023benchmark}, using the pre-trained TransOCR~\cite{chen2021scene,yu2021benchmarking} model. The evaluation is conducted on the real-world dataset RealCE~\cite{ma2023benchmark}.
As shown in Table~\ref{comparison4}, PTDB methods significantly improve performance on both tasks when combined with our latent refinement and interaction strategy.

\begin{table*}[t]
    \footnotesize
    \centering
\caption{The comparison for image deraining results. }
\label{table:deraining}
\begin{tabular}{|l| c c| c c| c c|c c| c c| c c|}
\hline
 \textbf{Method} & PSNR~$\textcolor{black}{\uparrow}$ & SSIM~$\textcolor{black}{\uparrow}$ & PSNR~$\textcolor{black}{\uparrow}$ & SSIM~$\textcolor{black}{\uparrow}$ & PSNR~$\textcolor{black}{\uparrow}$ & SSIM~$\textcolor{black}{\uparrow}$ & PSNR~$\textcolor{black}{\uparrow}$ & SSIM~$\textcolor{black}{\uparrow}$ & PSNR~$\textcolor{black}{\uparrow}$ & SSIM~$\textcolor{black}{\uparrow}$ & PSNR~$\textcolor{black}{\uparrow}$ & SSIM~$\textcolor{black}{\uparrow}$ \\
\hline \hline
  & \multicolumn{2}{c|}{\textbf{Test100}}&\multicolumn{2}{c|}{\textbf{Rain100H}}&\multicolumn{2}{c|}{\textbf{Rain100L}}& \multicolumn{2}{c|}{\textbf{Test2800}}&\multicolumn{2}{c|}{\textbf{Test1200}}&\multicolumn{2}{c|}{\textbf{Mean Value}}\\ \hline
DiffBIR & {24.50} & {0.707} & {24.84} & {0.702} & {30.83} & {0.765} &27.75 & 0.734 & {27.26} &{0.721} & {27.04} & {0.726} \\
DiffBIR+Ours  &\textbf{25.06}  &\textbf{0.730}  &\textbf{25.75} &\textbf{0.741} &\textbf{31.40} &\textbf{0.793} &\textbf{29.03}  &\textbf{0.756} &\textbf{28.87}  &\textbf{0.750}  & \textbf{28.02} &\textbf{0.754}  \\ \hline
PASD& {25.30} & {0.718} & {25.59} & {0.723} & {31.17} & {0.776} & {28.79} & {0.748} & {28.54} & {0.737} & {27.88} & {0.740}  \\
PASD+Ours & \textbf{26.12} &\textbf{0.746}  &\textbf{26.47} &\textbf{0.758} &\textbf{32.15}&\textbf{0.801}&\textbf{29.56}  & \textbf{0.761}&\textbf{29.19}  & \textbf{0.752} & \textbf{28.70} &\textbf{0.764} \\
\hline
\end{tabular}
\vspace{-0.1in}
\end{table*}

\begin{table*}[t]
    \footnotesize
    \centering
    \caption{Single-image motion deblurring results. }
    \label{tab:motion}
    \begin{tabular}{|l|cc|cc|cc|cc|}
        \hline
        \multirow{2}{1cm}{\textbf{Method}} & \multicolumn{2}{c|}{\textbf{GoPro}} & \multicolumn{2}{c|}{\textbf{HIDE}} & \multicolumn{2}{c|}{\textbf{RealBlur-R}} & \multicolumn{2}{c|}{\textbf{\textbf{RealBlur-J}}} \\
        & PSNR$\textcolor{black}{\uparrow}$ & {SSIM}$\textcolor{black}{\uparrow}$ & PSNR$\textcolor{black}{\uparrow}$ & {SSIM}$\textcolor{black}{\uparrow}$ & PSNR$\textcolor{black}{\uparrow}$ & {SSIM}$\textcolor{black}{\uparrow}$ & PSNR$\textcolor{black}{\uparrow}$ &{SSIM}$\textcolor{black}{\uparrow}$\\
        \hline \hline
        DiffBIR & {27.99} & {0.862} &	{26.88} &{{0.844}} & {31.87} &  {0.853} & 24.53& {0.772} \\
        +Ours & \textbf{28.73}&\textbf{0.893} &\textbf{27.59}  &\textbf{0.876} &\textbf{32.71}  &\textbf{0.878} &\textbf{25.87} &\textbf{0.781}\\ \hline
        PASD & {28.67} & {{0.881}} & {27.41} & {{0.858}} & {32.64} & {{0.869}} & {25.36} & {{0.793}}\\
        +Ours & \textbf{29.51}& \textbf{0.914}&\textbf{28.84}  & \textbf{0.883}&\textbf{33.32}  &\textbf{0.891} & \textbf{26.54}&\textbf{0.815}\\
        \hline
    \end{tabular}
\end{table*}

\begin{table*}[t]
    \footnotesize
    \centering
    \caption{Defocus deblurring comparisons on the DPDD testset (containing 37 indoor and 39 outdoor scenes). \textbf{S:} single-image defocus deblurring. \textbf{D:} dual-pixel defocus deblurring. }
    \label{table:dpdeblurring}
    \begin{tabular}{|l|cc|cc|cc|cc|}
\hline
   \multirow{2}{1cm}{\textbf{Method}} & \multicolumn{2}{c|}{\textbf{Indoor Scenes} (S)} & \multicolumn{2}{c|}{\textbf{Outdoor Scenes} (S)} & \multicolumn{2}{c|}{\textbf{Indoor Scenes} (D)} & \multicolumn{2}{c|}{\textbf{Outdoor Scenes} (D)}  \\
\cline{2-9}
    & PSNR$\textcolor{black}{\uparrow}$ & SSIM$\textcolor{black}{\uparrow}$ & PSNR$\textcolor{black}{\uparrow}$ & SSIM$\textcolor{black}{\uparrow}$ & PSNR$\textcolor{black}{\uparrow}$ & SSIM$\textcolor{black}{\uparrow}$& PSNR$\textcolor{black}{\uparrow}$ & SSIM$\textcolor{black}{\uparrow}$  \\
\hline \hline
{DiffBIR}& {25.34}  & {0.808}  & {20.07}  & {0.663}   & {25.91}  & {0.835}   & {20.67}  & {0.684}     \\
{DiffBIR}+Ours & \textbf{26.47}&\textbf{0.812}  &\textbf{21.58} &\textbf{0.681} &\textbf{26.75} &\textbf{0.847} &\textbf{21.42} &\textbf{0.701}    \\
PASD &27.36 & 0.844  & 20.75 &0.702 & 27.02& 0.863 & 21.90& 0.708\\
PASD +Ours &\textbf{28.29} &\textbf{0.855}  &\textbf{21.33} &\textbf{0.720} &\textbf{27.87} &\textbf{0.879} &\textbf{22.26} &\textbf{0.725}  \\
\hline
\end{tabular}
\end{table*}

\subsection{Evaluation with General Restoration}
\label{sec:sr-other}
We find that our strategy is applicable to tasks beyond low-light image enhancement. 

\noindent\textbf{SR.}
First, we conduct experiments on SR datasets. Following the experimental settings of DiffBIR~\cite{lin2024diffbir}, we evaluate performance using PSNR, SSIM, LPIPS~\cite{zhang2018unreasonable}, MANIQA~\cite{yang2022maniqa}, MUSIQ~\cite{ke2021musiq}, and CLIP-IQA~\cite{wang2023exploring}. The test set includes the synthetic dataset DIV2K-Val~\cite{agustsson2017ntire} and the real-world dataset RealSRSet~\cite{zhang2021designing}. As shown in Table~\ref{comparison1-sr} and \ref{comparison2-sr}, our method improves pre-trained diffusion-based approaches for SR tasks.

\noindent\textbf{Other Tasks.}
We further select key tasks such as deraining, motion deblurring, and defocus deblurring for evaluation.
For deraining, we use the Rain13K~\cite{zamir2022restormer} dataset for training and evaluate on the Rain100H~\cite{yang2017deep}, Rain100L~\cite{yang2017deep}, Test100~\cite{zhang2019image}, Test1200~\cite{zhang2018density}, and Test2800~\cite{fu2017removing} datasets. For single-image motion deblurring, we train on the GoPro~\cite{gopro2017} dataset and evaluate on synthetic datasets (GoPro~\cite{gopro2017}, HIDE~\cite{shen2019human}) as well as real-world datasets (RealBlur-R~\cite{rim_2020_realblur}, RealBlur-J~\cite{rim_2020_realblur}). For defocus deblurring, we use the DPDD~\cite{abdullah2020dpdd} training data and test on the EBDB~\cite{karaali2017edge_EBDB} and JNB~\cite{shi2015just_jnb} datasets. The pre-trained restoration model is Restormer.

The experimental results for the deraining task are shown in Table~\ref{table:deraining}, for motion deblurring in Table~\ref{tab:motion}, and for defocus deblurring in Table~\ref{table:dpdeblurring}. As observed, our method consistently improves performance across all three tasks, with non-trivial gains. In the future, we plan to explore the potential of our method in a wider range of tasks.

\noindent\textbf{Clarification.}
\textit{Note that the primary focus of this paper is the low-light enhancement task, owing to its representative difficulty and strong practical relevance. The experiments on other tasks are mainly conducted to demonstrate the potential of our method in broader image restoration scenarios, further highlighting its effectiveness and underlying insights. More extensive exploration of these tasks will be carried out in future work.}

%% file: sec/5_conclusion.tex
\section{Conclusion}
In this paper, we propose a plug-and-play approach for conditional latent modeling in low-light image enhancement using pre-trained diffusion models. We introduce a novel method that generates an appropriate generative prior for latent refinement and then predicts the refined latent with high fidelity. Additionally, we highlight the benefits of allowing the refined latent condition to dynamically interact with the noisy latent, leading to improved restoration performance.
Extensive experiments on various datasets demonstrate significant fidelity improvements in PTDB methods.

In this work, the latent refinement and bidirectional interaction strategies demonstrate significant effectiveness. However, they also increase training and inference costs. In the future, we aim to develop more lightweight strategies for various diffusion models. Additionally, we aim to develop a unified algorithm for restoring diverse tasks, building on some promising results that have been demonstrated in this paper.